\documentclass[10pt,twocolumn,letterpaper]{article}

\usepackage{cvpr}              %

\usepackage{siunitx}
\usepackage{multirow,bigdelim}
\usepackage{amsmath,amssymb} %
\usepackage{color}
\usepackage{pbox}
\usepackage{wrapfig}
\usepackage{csquotes}
\usepackage{booktabs}
\usepackage{pifont}
\usepackage{makecell}
\usepackage{tabularx}
\usepackage{rotating}

\usepackage[sectionbib]{chapterbib}

\usepackage[dvipsnames]{xcolor}

\usepackage{mathtools}

\DeclarePairedDelimiter\parenstmp{\lparen}{\rparen}

\DeclarePairedDelimiter\normtmp{\lVert}{\rVert}

\DeclarePairedDelimiter\bracketstmp{\lbrack}{\rbrack}

\newcommand{\parens}[1]{\parenstmp*{#1}}

\newcommand{\norm}[1]{\normtmp*{#1}}

\newcommand{\brackets}[1]{\bracketstmp*{#1}}

\newcommand{\operatorcall}[2]{\operatorname{#1}\parens{#2}}
\newcommand{\R}{\mathbb{R}}

\newcommand{\E}{\mathbb{E}}

\newcommand{\gauss}{\mathcal{N}\left(\mathbf{0},\mathbf{I}\right)}
\newcommand{\noise}{\mathbf{n}}

\newcommand{\cmark}{\ding{51}}
\newcommand{\xmark}{\ding{55}}

\newcommand{\setofposes}{\R^{J \times 3}}
\newcommand{\pose}{\mathbf{s}}
\newcommand{\image}{\mathbf{x}}
\newcommand{\context}{\mathbf{z}}

\newcommand{\controlnet}{CN}

\newcommand{\depth}{\mathbf{c}_d}
\newcommand{\keypoints}{\mathbf{c}_k}
\newcommand{\semantics}{\mathbf{c}_s}
\usepackage{microtype}
\usepackage{array}

\newcommand{\z}{\phantom{0}}
\newcommand{\suppmat}{Supp.\@ Mat.\@}

\definecolor{tabblue}{HTML}{1f77b4}
\definecolor{taborange}{HTML}{ff7f0e}
\definecolor{tabgreen}{HTML}{2ca02c}
\definecolor{tabred}{HTML}{d62728}
\definecolor{tabpurple}{HTML}{9467bd}
\definecolor{tabbrown}{HTML}{8c564b}
\definecolor{tabpink}{HTML}{e377c2}
\definecolor{tabgray}{HTML}{7f7f7f}
\definecolor{tabolive}{HTML}{bcbd22}
\definecolor{tabcyan}{HTML}{17becf}

\setlist[itemize]{noitemsep,leftmargin=*,topsep=0em}
\setlist[enumerate]{noitemsep,leftmargin=*,topsep=0em}

\usepackage{comment}

\definecolor{cvprblue}{rgb}{0.21,0.49,0.74}
\newcommand\attrhighlight[1]{\textbf{\textcolor{cvprblue}{#1}}}
\usepackage[pagebackref,breaklinks,colorlinks,citecolor=cvprblue]{hyperref}
\usepackage[capitalize]{cleveref}
\crefname{section}{Sec.}{Secs.}
\Crefname{section}{Section}{Sections}
\Crefname{table}{Table}{Tables}
\crefname{table}{Tab.}{Tabs.}
\def\paperID{334} %
\def\confName{3DV\xspace}
\def\confYear{2026\xspace}

\newcommand{\mytitle}{\emph{Are Pose Estimators Ready for the Open World?} \\ STAGE: A GenAI Toolkit for Auditing 3D Human Pose Estimators}
\title{\vspace{-3em} \mytitle}
\supptitle{\mytitle}

\author{Nikita Kister\textsuperscript{1}%
\quad
István Sárándi\textsuperscript{2}%
\quad
Jiayi Wang\textsuperscript{3}%
\quad
Anna Khoreva\textsuperscript{4}
\quad
Gerard Pons-Moll\textsuperscript{2,5}\\[5pt]
{\small
\textsuperscript{1}Bosch IoC Lab, University of Tübingen \quad
\textsuperscript{2}Tübingen AI Center, University of Tübingen \quad
\textsuperscript{3}Bosch Center for Artificial Intelligence}\\
{\small \textsuperscript{4}Zalando SE \quad
\textsuperscript{5}Max Planck Institute for Informatics, Saarland Informatics Campus}\\[5pt]
{\tt\small \href{https://virtualhumans.mpi-inf.mpg.de/stage/}{https://virtualhumans.mpi-inf.mpg.de/stage/}}}

\renewcommand{\paragraph}[2][\ ]{\vspace{6pt}\noindent{\bf #2.}}

\begin{document}

\twocolumn[{%
\renewcommand\twocolumn[1][]{#1}%
\maketitle%
\newcommand\promptstyle[1]{\textit{\enquote{#1}}}%
\centering
\captionsetup{type=figure}
\small
\vspace{-2em}
\begin{tabular}{@{\hspace{-0.5em}}c@{\hskip 3pt}c@{\hskip 8pt}c@{\hskip 3pt}c@{\hskip 8pt}c@{\hskip 3pt}c@{\hskip 0pt}c@{}}
\multicolumn{2}{c}{\pbox{0.19\textwidth}{\makecell{\makebox[0pt][c]{Base image and predicted pose}}}} &
\multicolumn{4}{c}{\pbox{0.56\textwidth}{\attrhighlight{Attribute} images and uncovered failures}} &
\multicolumn{1}{c}{\pbox{0.13\textwidth}{\hfill}} \\
\multicolumn{2}{c}{\pbox{0.28\textwidth}{\promptstyle{young man in gym}}} &
\multicolumn{2}{c}{\pbox{0.28\textwidth}{\promptstyle{young \attrhighlight{woman} in gym}}} &
\multicolumn{2}{c}{\pbox{0.28\textwidth}{\promptstyle{\attrhighlight{elderly} man in gym}}} &
\multicolumn{1}{c}{\pbox{0.13\textwidth}{\centering \phantom{m} GT pose}} \\
\includegraphics[width=0.14\textwidth]{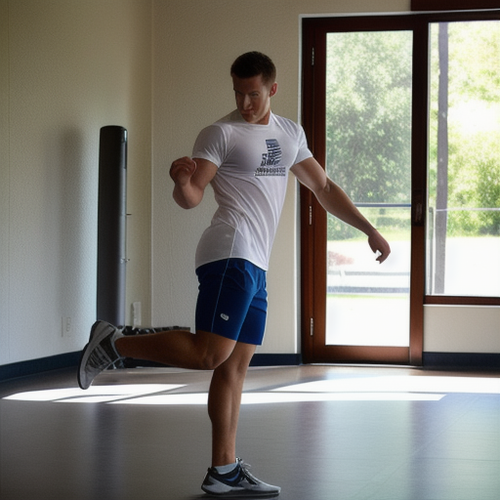} & \includegraphics[width=0.14\textwidth]{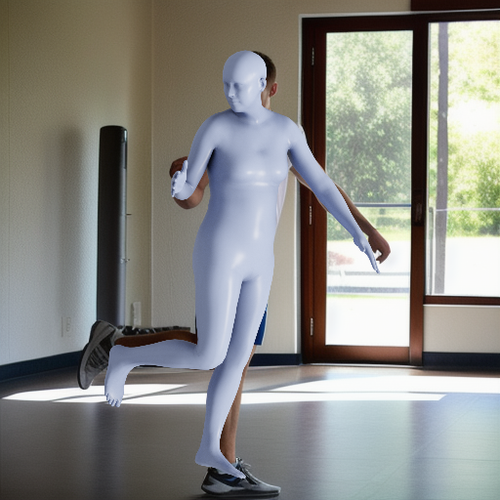} & \includegraphics[width=0.14\textwidth]{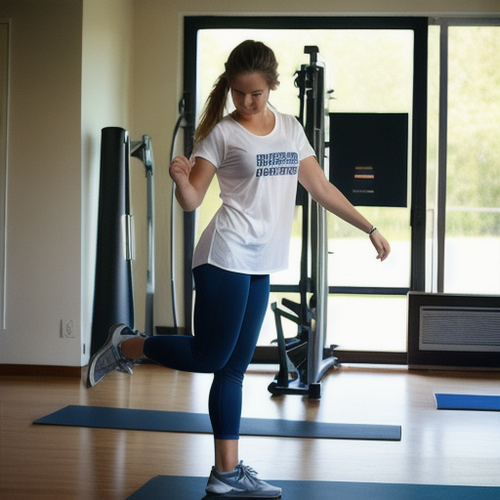} & \includegraphics[width=0.14\textwidth]{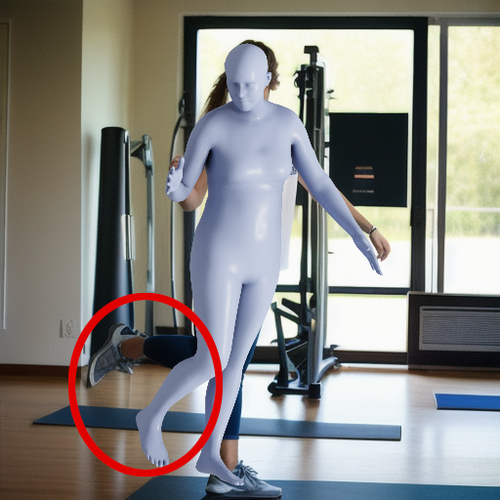} & \includegraphics[width=0.14\textwidth]{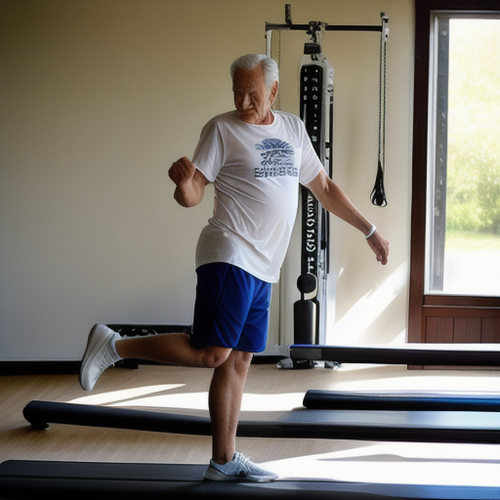} & \includegraphics[width=0.14\textwidth]{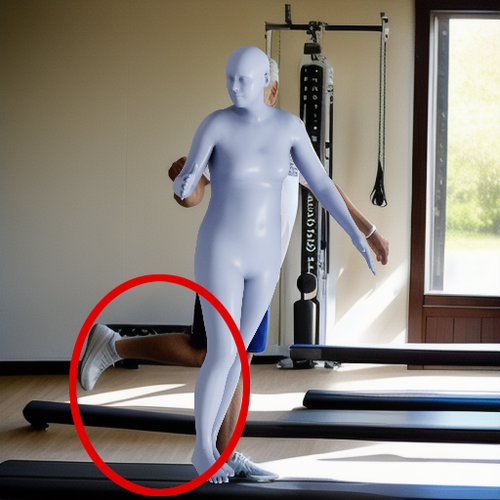} & \hspace{-0.5em} \includegraphics[width=0.14\textwidth]{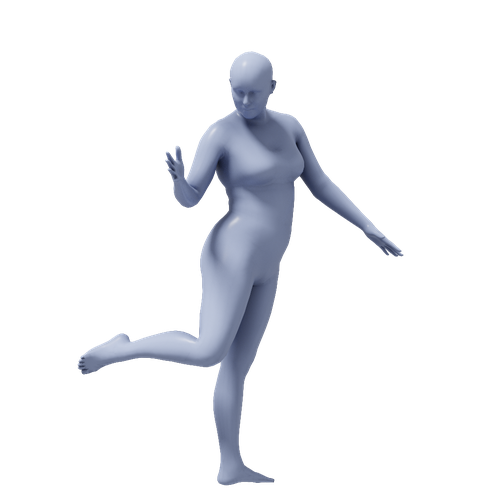} \hspace{-2.5em}  \\
\multicolumn{2}{c}{\pbox{0.28\textwidth}{\promptstyle{man wearing t-shirt in city}}} &
\multicolumn{2}{c}{\pbox{0.28\textwidth}{\promptstyle{man wearing \attrhighlight{floral shirt} in city}}} &
\multicolumn{2}{c}{\pbox{0.28\textwidth}{\promptstyle{man wearing \attrhighlight{parka} in city}}} &
\multicolumn{1}{c}{\pbox{0.13\textwidth}{\hfill}} \\
\includegraphics[width=0.14\textwidth]{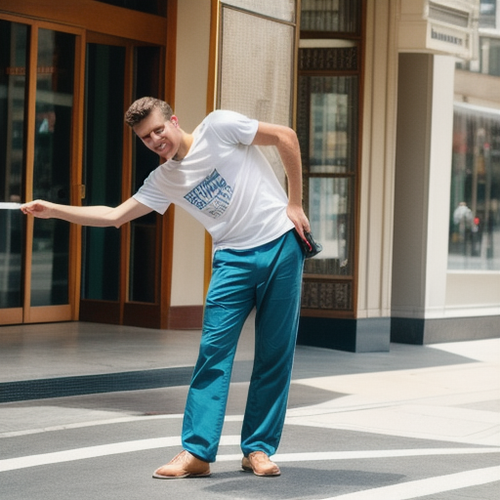} & \includegraphics[width=0.14\textwidth]{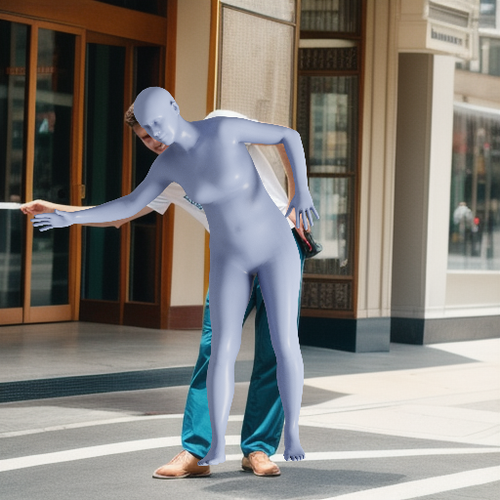} & \includegraphics[width=0.14\textwidth]{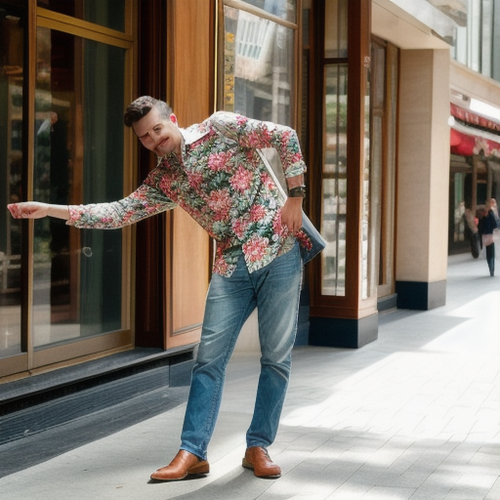} & \includegraphics[width=0.14\textwidth]{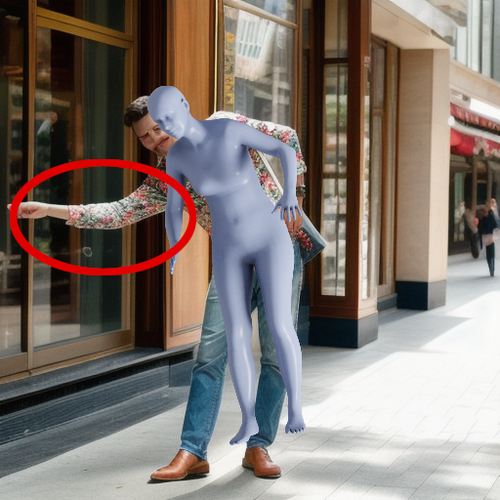} & \includegraphics[width=0.14\textwidth]{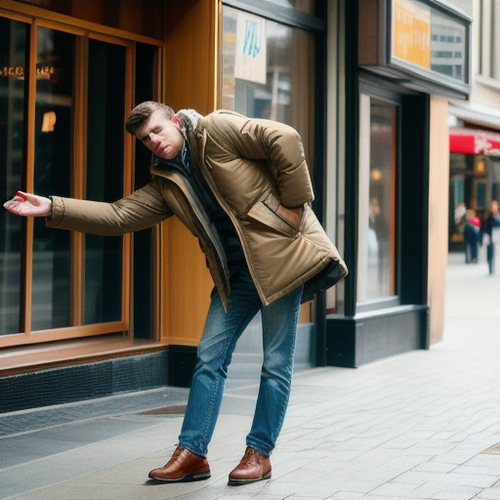} & \includegraphics[width=0.14\textwidth]{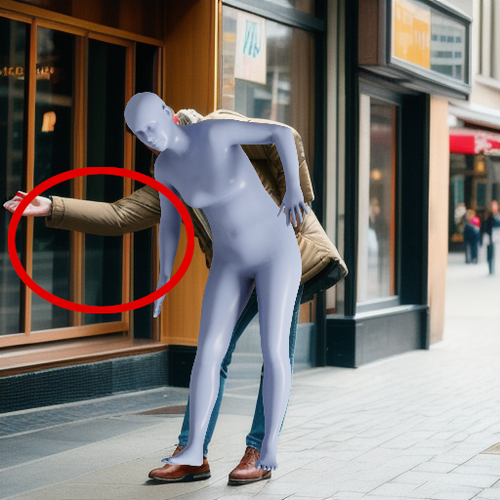} & \includegraphics[width=0.14\textwidth]{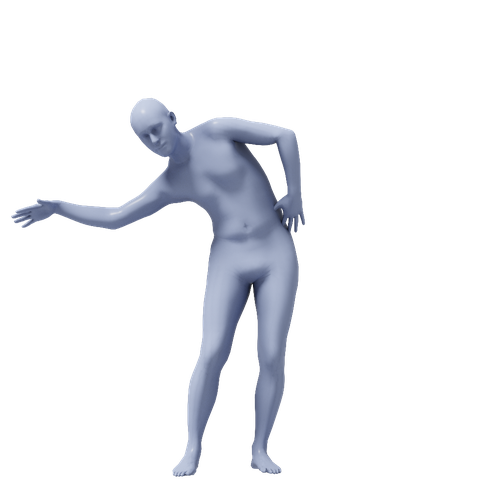} \hspace{-3em}   \\
\end{tabular}
\vspace{-0.8em}
\captionof{figure}{
\textbf{How robust are 3D human pose estimators to different attributes such as gender, age, or clothing?}
To measure this, \emph{STAGE} generates a base and an attribute image, where the difference is one specific control attribute---the pose is the same, and all other attributes are kept similar. In contrast to real benchmarks, this allows us to isolate the effect of one attribute on the pose estimator's performance. E.g., notice how changing the gender (row 1) or clothing (rows 2) can lead to significant pose estimation failures.}
\label{fig:teaser}
\vspace{1.5em}
}]
\begin{abstract}%
\noindent%
For safety-critical applications, it is crucial to audit 3D human pose estimators before deployment.
Will the system break down if the weather or the clothing changes? Is it robust regarding gender and age?
To answer these questions and more, we need controlled studies with images that differ in a single attribute, but real benchmarks cannot provide such pairs.
We thus present STAGE, a GenAI data toolkit for auditing 3D human pose estimators.
For STAGE, we develop the first GenAI image creator with accurate 3D pose control and propose a novel evaluation strategy to isolate and quantify the effects of single factors such as gender, ethnicity, age, clothing, location, and weather.
Enabled by STAGE, we generate a series of benchmarks to audit, for the first time,  the sensitivity of popular pose estimators towards such factors. 
Our results show that natural variations can severely degrade pose estimator performance, raising doubts about their readiness for open-world deployment. We aim to highlight these robustness issues and establish STAGE as a benchmark to quantify them.
\end{abstract}
\section{Introduction}

3D human pose estimation (HPE) is an important component of safety-critical systems that operate among humans, such as autonomous vehicles and collaborative robots.
It is crucial to ensure that models are robust towards elements naturally occurring in the operational domain.
Coarse-grained measures like average error over a test set are not sufficient, we need targeted analysis regarding individual attributes such as weather, location, gender, age, etc.
The need for such evaluations has also been recognized by legislators in the recent EU AI Act~\cite{euai} and the UL 4600 safety standard for evaluation for autonomous products~\cite{ul}.

In most fields of science, controlled experiments are the standard way to isolate the effects of one variable at a time.
To perform such experiments for robustness analysis in HPE, one would need pairs of images that differ only by a single attribute, but this is not available at present.
Benchmarks consisting of real images \cite{h36m_pami,3dhp,3dpw,emdb} cannot provide such control---it is infeasible to capture a sufficient variety of people in the same exact pose while varying attributes. Such benchmarks are also laborious to capture, and the attributes of interest vary depending on application, requiring repeated data acquisition.

Synthetic benchmarks generated with computer graphics (CG)~\cite{surreal, agora, gta, bedlam, synbody, hspace} enable precise control over body pose, camera position, clothing, etc., but are time-consuming and require an abundance of high-quality 3D assets to achieve photorealism and diversity.
Clothing has to be designed and simulated~\cite{bedlam}, and scenes have to be modeled or captured.
Effective use of modern graphics pipelines also requires expertise that HPE practitioners may lack.
Overall, the current tools do not allow for a thorough auditing of HPE models before their deployment in the open world.

With the recent rise of high-quality image generators, we have the opportunity to \emph{synthesize} the required image pairs, precisely controlling attributes by simply conditioning on text and pose.
For this to be viable, the generated images must be 1) sufficiently realistic and 2) coherent with the specified 3D pose and text prompt.
Existing conditional image generators such as ControlNet (CN)~\cite{controlnet} do not satisfy such requirements.
At best, recent methods allow control over 2D pose \cite{controlnet, t2i, humansd}, which is insufficient, as several 3D poses can project to the same 2D pose~\cite{taylorcamillo,3d2dambiguity}. 
Depth map-conditioned models~\cite{controlnet} are not suitable either---they cannot produce complex backgrounds and complex clothing when given human depth maps only.

Our first contribution is, therefore, an improved human image generator with precise 3D pose control in the SMPL format~\cite{smpl}.
Our model synthesizes high-quality images that are also consistent with the conditioning 3D pose, making it suitable for evaluation.
We experimentally show that a pretrained SOTA 3D pose estimator can infer the intended pose from our generated images with similar accuracy as from real images, validating pose coherence.
To further verify data quality, we also use the generated images for training.
Recent work has shown SOTA pose estimation results by training purely on CG images~\cite{bedlam}, though this required many hours of engineering work and computing resources.
Remarkably, we achieve the same performance with our fully automatic GenAI toolkit, where diversity scales with simple text prompts instead of novel asset design as with CG.

\begin{figure*}[tp]
\vspace{-0.9em}
\centering
    \includegraphics[width=\textwidth]{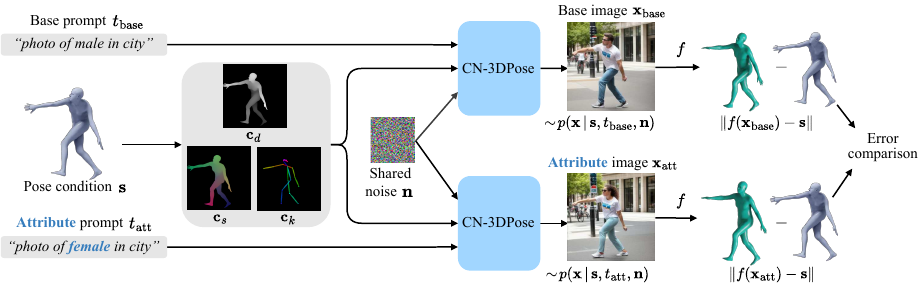}
    \vspace{-2em}
    \caption{\textbf{Evaluating estimators with STAGE.}
    Given base and attribute prompts along with a desired 3D pose condition, we generate a pair of test images using our CN-3DPose image generator.
    We encode the pose as a depth map, a dense semantic encoding and a 2D skeleton drawing.
    To preserve visual similarity between the images except for the target attribute, we use the same initial noise for both images.
    The result is a test image pair that differs mostly by our attribute of interest, suitable for controlled experiments.
    We run the pose estimator to be tested on both images and compare the prediction errors to measure its sensitivity towards the attribute.
    }
    \label{fig:model:architecture}
    \vspace{-0.5em}
\end{figure*}

Our goal in this work is to enable the easy generation of diverse benchmarks for controlled studies to audit 3D pose estimators \wrt single attributes.
We achieve this for the first time with our proposed \emph{STAGE} framework, which encompasses the aforementioned 3D-controllable generator and an evaluation methodology with appropriate metrics and protocols.
STAGE allows the user to specify any target attributes in text form, \eg a clothing item such as \enquote{parka}.
From this, it generates two sets of synthetic images, one where people wear parka coats and one where they do not, while keeping the appearance of the images as close as possible in other aspects, see \cref{fig:teaser}.
The performance difference between these two synthetic sets indicates the sensitivity of the pose estimator \wrt the attribute, which is then shown to the user.

STAGE can shed new light on aspects of SOTA model performance that can not be evaluated in current benchmarks.
To demonstrate its usefulness, we use STAGE to thoroughly audit popular SOTA HPE methods~\cite{pare, metrabs_acae, 4dhumans, bedlam, smplerx, pymafx, spin} for attribute categories such as clothing, location, weather, and protected attributes (gender, age, \etc), making several interesting findings.
For example, the best-performing methods, on average, are not always the most robust to attributes.
We will release our code and data for future research.
In summary, our contributions are:
\begin{itemize}
    \item We propose STAGE, a customizable GenAI benchmark generator, along with a systematic evaluation protocol that allows, for the first time, to conduct controlled experiments to audit 3D human pose estimators. 
    \item Using STAGE, we audit popular pose estimators on their sensitivity against attributes such as gender, age, clothing, location, making findings that were not possible before.
    \item We build a text-to-image generative model that can create images with coherent 3D pose. The quality is on par with the best CG-rendered datasets while being significantly simpler and cheaper to operate and scale.
\end{itemize}
\section{Related work}
\paragraph{3D human pose benchmarks}
High-quality benchmarks have been a major driving force of progress in HPE\@.
Real-image benchmarks \cite{h36m_pami, 3dhp, 3dpw, emdb, rich} are the gold standard but are expensive to capture at scale.
They are either restricted to a studio or have a limited number of subjects. High variation of clothing, locations, gender, and ethnicity has not been possible to achieve with real data.
Therefore, graphics-based synthetic benchmarks \cite{agora, gta, bedlam, synbody, surreal, hspace} have risen in popularity, allowing the generation of accurately annotated data at scale.
However, they require laborious asset design (clothes, 3D environments) as well as expertise to operate complex graphics, animation, and physics simulation pipelines.
Thus, even graphics-based benchmarks are limited in terms of diverse factors that can impact performance.

\paragraph[]{GenAI for human images} 
Methods can be categorized by whether they use 2D or 3D pose control.
2D-pose-conditioned methods \cite{controlnet, controlnetxs, controlnet_plus_plus, t2i, humansd, hyperhuman, hallu, envpeople}  extend StableDiffusion (SD)~\cite{ldm} with 2D keypoint control.
They exhibit good text control but do not allow precise control of human pose in 3D.
While ControlNet \cite{controlnet} can combine multiple control signals (such as depth and 2D pose), the resulting model achieves only limited 3D controllability, as we show in \cref{sec:exp:quality}.
For 3D pose control methods, we can broadly speak of GAN-based and score distillation--based methods.
GAN-based methods \cite{enarf, humangan, eva, gnarf, ag3d, warp3dreposing, stylepeople} learn to generate 3D humans from 2D single-view image collections.
Through techniques such as inverse skinning, they achieve excellent pose control.
However, they generate people without backgrounds and do not offer text control.
Some methods based on score distillation sampling \cite{dreamfusion, dreamhuman, dreamavatar, tada} or CLIP \cite{clip, clipactor,avatarclip} offer text control.
However, their image quality is often insufficient, and generation can take hours.
Finally, while some prior works \cite{diffusionhpc, humanwild, posesyn} have used human image generation to supplement HPE training data, we are the first to leverage it for evaluating HPE robustness.

\paragraph{Auditing with GenAI}
The utility of GenAI models for benchmarking and auditing has been shown before for object classification and detection \cite{domino, promptattack, bugs, testing, scrod, imagenete, counterfact1, counterfact2, zeroshot, lance, datasetinterface, facebiasbench, morphgan}.
Some methods search for a semantically coherent subspace of images that cause prediction failure \cite{domino, promptattack, bugs, testing, scrod}, while others create counterfactual images to measure classifier robustness \cite{counterfact2, counterfact1, zeroshot, lance, datasetinterface}.
Unlike these classification and detection works, we are targeting HPE, a more fine-grained, high-dimensional regression problem requiring a more elaborate approach.
\section{Method}

We propose STAGE, a synthetic data generation toolkit to create custom benchmarks for auditing HPE methods\@.
In \cref{sec:problem} we introduce the problem formulation and our evaluation protocol.
We describe our 3D pose-conditioned image generator in \cref{sec:model} and finish with the benchmark generation in \cref{sec:datagen}.

\subsection{Problem formulation}
\label{sec:problem}
A 3D pose estimator $f: \image \mapsto \hat{\pose}$ maps an image of a person $ \image \in \mathbb{R}^{H\times W \times 3}$ to a skeleton consisting of $J$ joint locations $\hat{\pose}\in\setofposes$.
Pose estimators are deployed in some operational domain where image $\image$ and pose $\pose$ follow a distribution $p(\image, \pose)$. They are typically evaluated with a metric $L(\hat{\pose}, \pose)$ by the risk %
\begin{equation}
  R(f) = \E_{\image, \pose} \brackets{L(f(\image), \pose)},
\label{eq:standard_risk}
\end{equation} 
where sampling from $p(\image, \pose)$ is approximated by iterating through a captured and annotated dataset.

\begin{figure*}[t]
\centering
\small
\vspace{-0.2em}
\begin{tabular}{ccccc}
\pbox{0.15\linewidth}{\centering 3D pose condition} & \pbox{0.15\linewidth}{\centering High BMI} & Floral texture & Trench coat & Wine cellar \\
\includegraphics[width=0.15\linewidth]{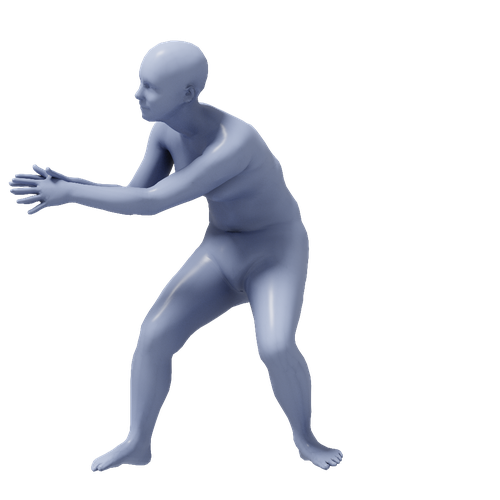} &%
\includegraphics[width=0.15\linewidth]{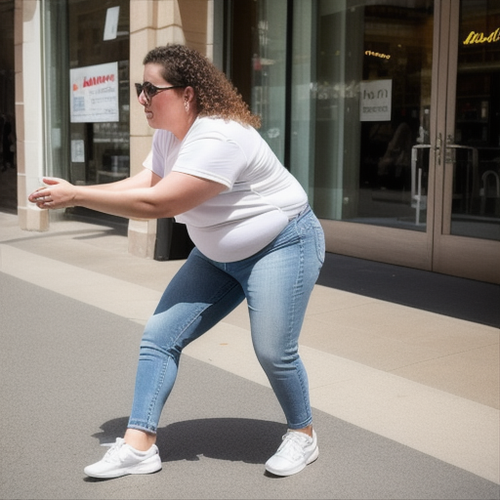} &%
\includegraphics[width=0.15\linewidth]{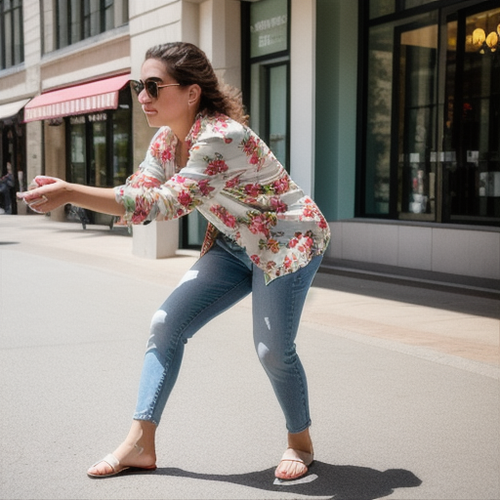} &%
\includegraphics[width=0.15\linewidth]{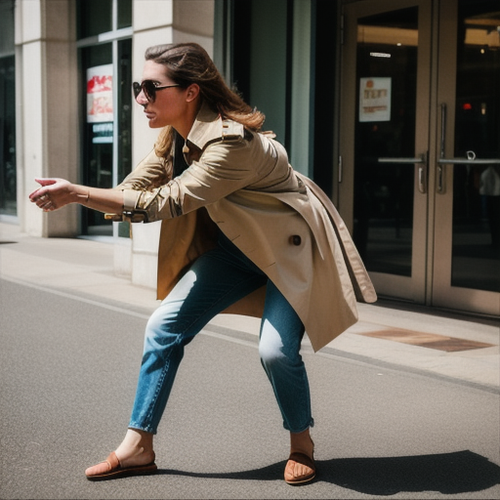} &%
\includegraphics[width=0.15\linewidth]{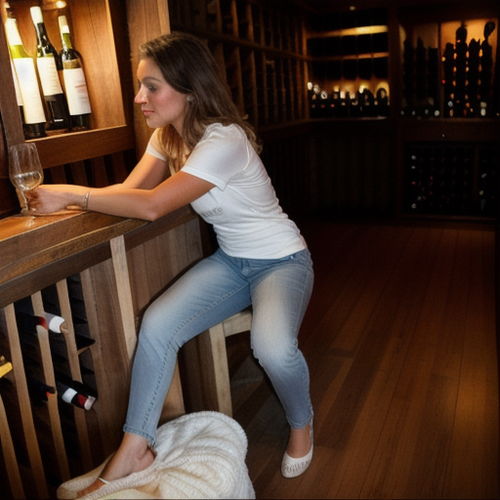} %
\\
\includegraphics[width=0.15\linewidth]{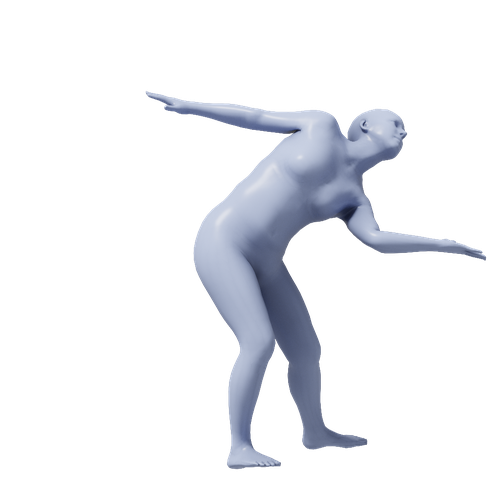} &%
\includegraphics[width=0.15\linewidth]{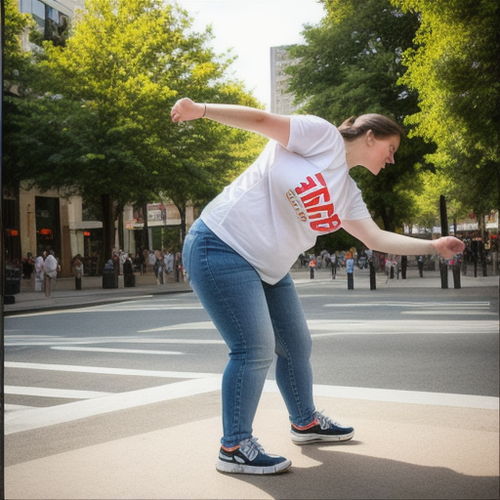} &%
\includegraphics[width=0.15\linewidth]{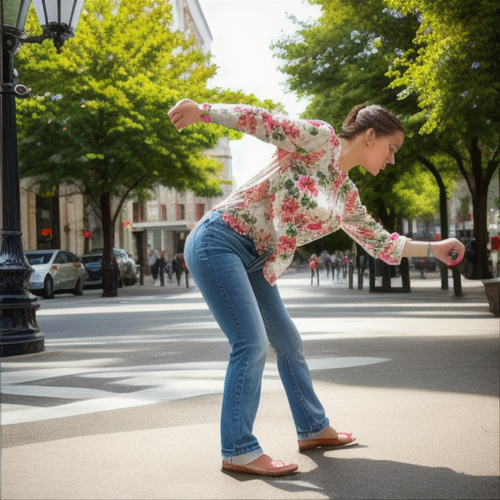} &%
\includegraphics[width=0.15\linewidth]{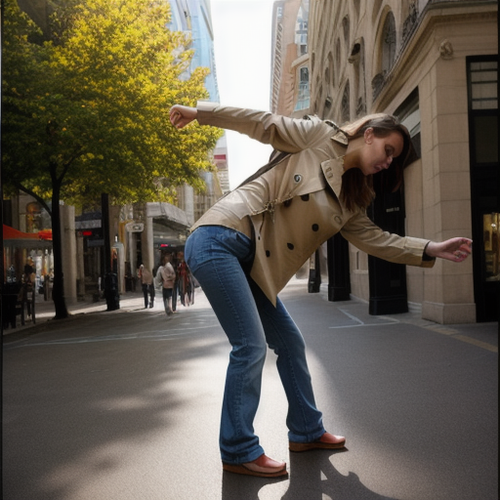} &%
\includegraphics[width=0.15\linewidth]{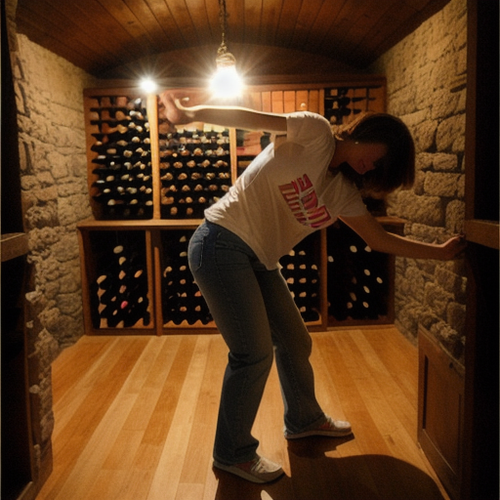} %
\\
\includegraphics[width=0.15\linewidth]{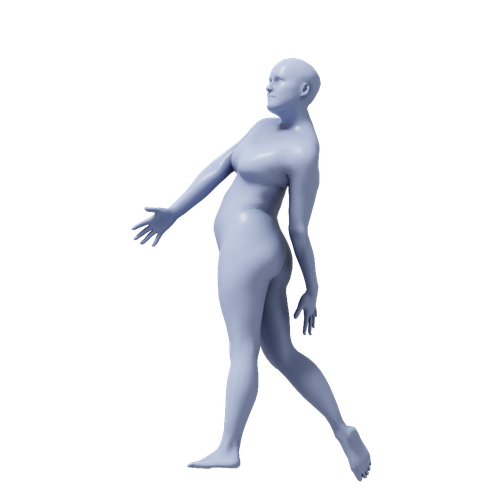} &%
\includegraphics[width=0.15\linewidth]{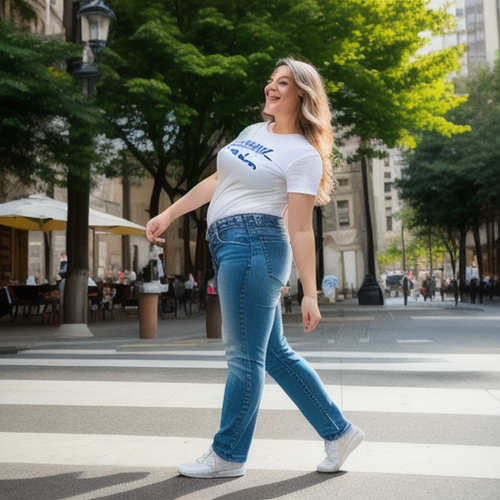} &%
\includegraphics[width=0.15\linewidth]{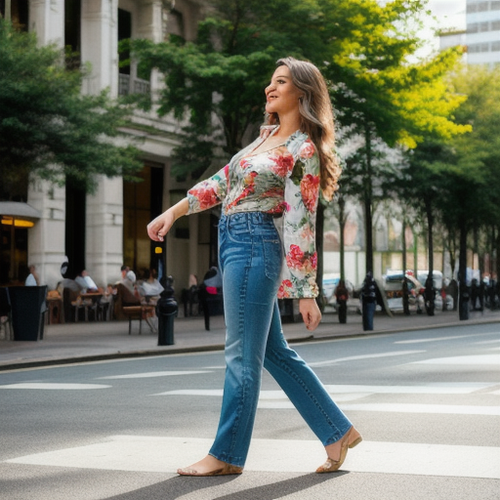} &%
\includegraphics[width=0.15\linewidth]{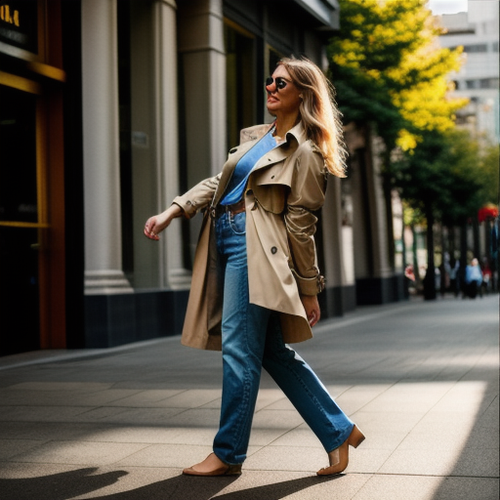} &%
\includegraphics[width=0.15\linewidth]{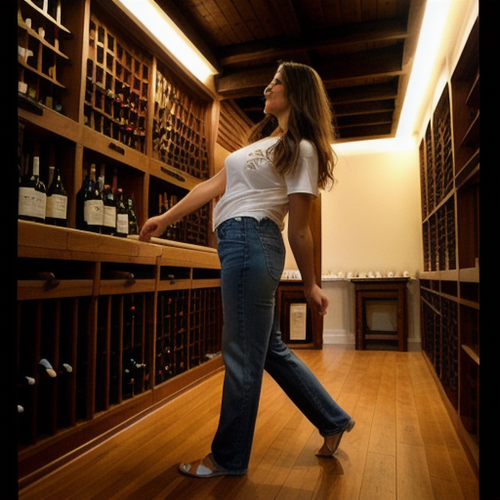} %
\\
\includegraphics[width=0.15\linewidth]{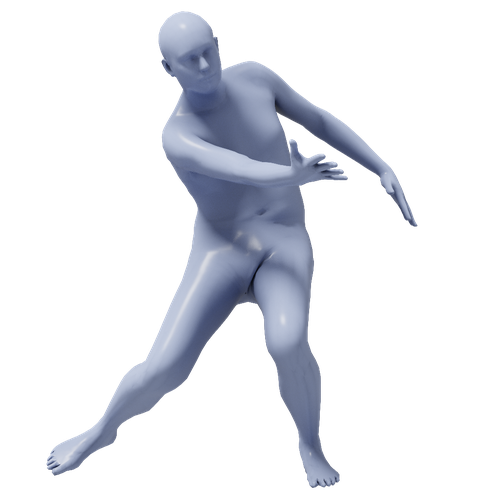} &%
\includegraphics[width=0.15\linewidth]{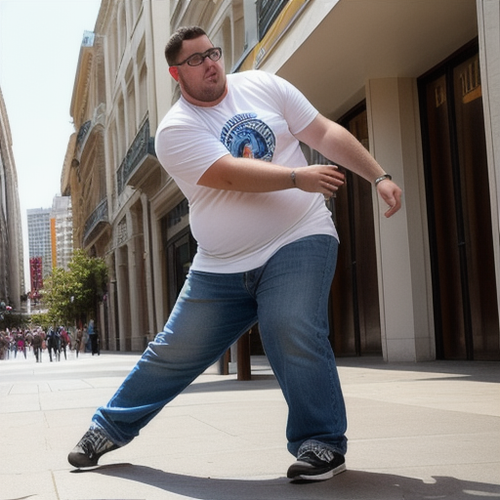} &%
\includegraphics[width=0.15\linewidth]{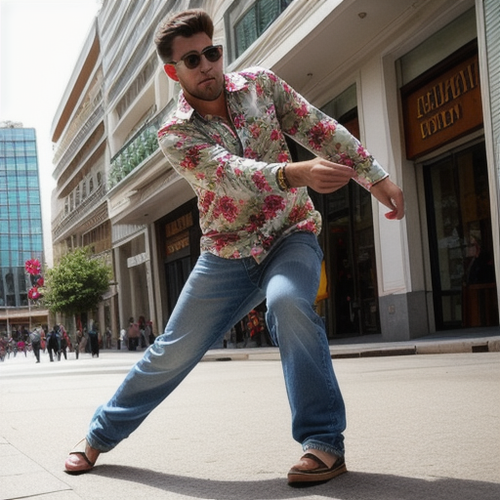} &%
\includegraphics[width=0.15\linewidth]{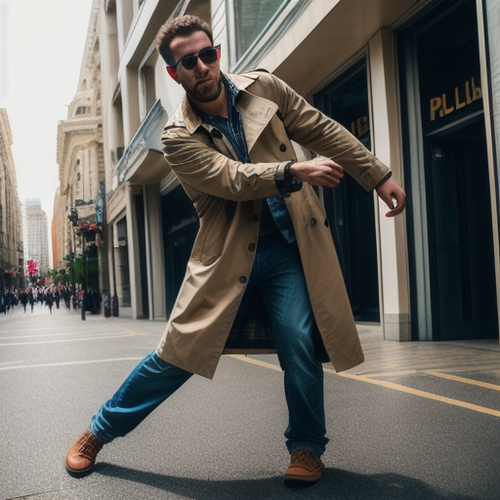} &%
\includegraphics[width=0.15\linewidth]{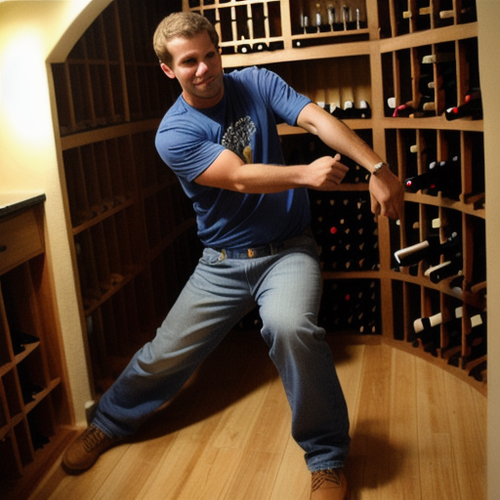} %
\\
\end{tabular}
\vspace{-0.5em}
\caption{\textbf{Images generated via STAGE.} 
By simply varying the text prompt, STAGE can generate images of people with different body shapes and appearances in different locations, while remaining well-aligned with the specified 3D pose.
When only the person's description is changed, the background remains similar.
This is important for our controlled experiments targeting single attributes.
}%
\label{fig:gallery}
\vspace{-1.0em}
\end{figure*}

For a given pose estimator, STAGE aims to isolate the effect of individual attributes such as clothing or location on its results.
Symbolically, we summarize such appearance details in a latent context $\context$.
Thus, for controlled experiments, we need to sample from $p(\image, \pose|\context)=p(\image| \pose,\context)p(\pose|\context)$.
We assume that pose is independent of attributes, \ie, $p(\pose|\context)=p(\pose)$.
For the prior $p(\pose)$ we use a large pose collection \cite{AMASS:ICCV:2019}, and approximate $\mathbf{x} \sim p(\image| \pose, \context)$ with a diffusion model as $\mathbf{x} = g(\pose, t, \noise)$, where $t$ is a text prompt and $\noise \sim \gauss$ is the initial noise for denoising. The model $g$ is our proposed CN-3DPose, described in \cref{sec:model}.
The context $\context$ is represented by $t$ and $\noise$.
We set the required attributes through text, \eg, specifying \emph{\enquote{photo, a caucasian elderly male wearing a shirt in the city at night.}}
The noise $\noise$ encodes appearance details not given in $t$, such as the precise background layout.

STAGE compares performance for \emph{two generated images at a time}, $\mathbf{x}_\text{base}=g(\pose, t_\text{base}, \noise)$ and $\mathbf{x}_\text{att}=g(\pose, t_\text{att}, \noise)$, where the prompts differ only by the attribute of interest while the initial noise is shared.
Using pairs of synthetic images reduces the estimation noise caused by imperfections in generation and computing the risk over many pairs reduces the estimation noise caused by random factors.
The base prompt $t_\text{base}$ describes a scene and human appearance that is well-represented in existing benchmarks, and should not be challenging for the pose estimator.
The images are fed to the pose estimator to get $\hat{\pose}_\text{base}=f(\mathbf{x}_\text{base})$ and $\hat{\pose}_\text{att}=f(\mathbf{x}_\text{att})$.

To measure sensitivity, we need to identify cases where the quality of $\hat{\pose}_\text{att}$ is \textit{degraded} compared to $\hat{\pose}_\text{base}$, \ie, if its error is worse than the base case by more than a given threshold $\tau$.
We count how often this happens in our proposed \emph{Percentage of Degraded Poses} metric $\operatorcall{PDP}{f, t_\text{base}, t_\text{att}}=$
\begin{equation}
     \frac{1}{N}\sum_{i=1}^N \brackets{L\parens{\hat{\pose}_\text{att}, \pose} > L\parens{\hat{\pose}_\text{base}, \pose} + \tau}.
\end{equation}
Note that the PDP is nonnegative by construction.
It aims to count the especially high-risk instances where a pose estimator performs well under a common benchmark scenario but degrades when moving to the open world, indicating non-robustness.

We consider it a problem if \emph{any} joint is significantly displaced, since this can change the pose semantics. 
Hence, for $L$ we use the Maximum Joint Error (MaxJE)---the highest joint error of a predicted pose.
For a ground-truth pose $\pose$, and predicted pose $\hat{\pose}$ (Procrustes-aligned \cite{procrustes}) with joints $\pose_i, \hat{\pose}_i\in\R^3$ respectively, the maximum joint error is
\begin{equation}
    L\parens{\hat{\pose}, \pose} = \mathrm{MaxJE}\parens{\hat{\pose}, \pose} = \max_{1\leq i \leq J} \norm{\hat{\pose}_i - \pose_i}.
\end{equation}

We employ Procrustes alignment to account for differences in camera angle and scale ambiguity between the pair of images.
To compute the PDP for a category, we average the PDP for each attribute in the category.
The overall PDP score for a pose estimator is the mean across all categories.

\subsection{3D pose-conditioned image generation}
\label{sec:model}
In the previous section, we assume the image generator $g(\pose, \noise, t)$ models sampling from $p(\image| \pose, \context)$ well. For this to be true, two requirements need to be fulfilled.
First, it must provide accurate \textit{3D pose alignment} between the condition and the generated images.
Second, the effect of the conditioning pose and text prompt has to be disentangled. 
That is, the pose should not influence the shape or clothing.

We observed that existing image generators from \citet{controlnet} (\controlnet) do not meet the desired requirements.
This is because the two suitable available models, CN-Pose and CN-Depth, can only condition on a 2D pose or a depth map respectively. 
Images sampled from \controlnet-Pose suffer from poor 3D pose alignment since multiple 3D poses can project to the same 2D pose.
We also observe frequent swaps between the left and right side of the body, \eg generating back views instead of front views.
\controlnet-Depth %
is conditioned on a depth map, thus providing better 3D pose alignment.
However, \controlnet-Depth was trained with full image depth maps, while we only want to condition on the human and to rely on text for our scene description.
By parameterizing the desired 3D pose using the SMPL body model $\theta \in \operatorcall{SO}{3}^{24}$, the depth map of only the human can be used in \controlnet-Depth.
However, this leads to an oversampling of bland, flat backgrounds and simple body-tight clothing.
Even the combination of \controlnet-Pose and \controlnet-Depth (which we denote as \controlnet-Multi) generates too simple backgrounds and clothing as shown in \cref{tab:baseline}, \cref{fig:oursvs:imgs} and in the Supp.\@ Mat.  
Next, we propose our novel solution that overcomes these limitations of existing human image generators.

\paragraph{CN-3DPose}
Our method, CN-3DPose, is built on the pre-trained \controlnet-Depth using three key insights to effectively leverage both 3D and 2D pose datasets.
First, we fine-tune on isolated SMPL-based body depth maps $\depth$ \emph{without} background or clothing depth information.
This forces the model to rely on the text and pose control signals for creating diverse backgrounds and appearances while being coherent with the 3D pose. %
Second, to avoid confusion between body parts, we use dense semantic encoding $\semantics$, a rendering of SMPL with each vertex RGB-colored according to its XYZ coordinates in canonical pose, in addition to the skeleton drawing $\keypoints$.
Third, to mitigate over-fitting to the 3D pose dataset backgrounds and their limited variation in human appearance and poses, we train using both 3D and more diverse 2D pose datasets, as detailed in \cref{sec:exp:impl}.
For 2D-annotated examples, we mask the 3D pose conditions ($\depth$ and $\semantics$).
The resulting model meets our requirements for diversity and 3D pose alignment.
Overall, it lets us sample from 
$
    p(\image | \pose, \context) \approx p(\image|t, \depth, \semantics, \keypoints, \noise).
$
The key steps of our method are depicted in \cref{fig:model:architecture}.
\subsection{Benchmark data generation}
\label{sec:datagen}

\paragraph{Attribute categories} To broadly evaluate the capabilities of pose estimators in the open world, we choose several groups of attributes, including clothing and location, as well as protected attributes such as ethnicity, age, and gender to uncover potential biases.
We further consider adverse lighting and weather conditions, such as night, snow, or rain---an issue well-recognized in the autonomous driving community \cite{fog,night,rain,robocar}, but not yet in the HPE community. %
This leads to the prompt template \textit{\enquote{photo, \{ethnicity\} \{age\} \{gender\} wearing \{clothing\} in \{location\} at \{lighting condition\} \{weather\}}}. 
For the base prompt, we describe a common scene in existing benchmarks, with the intention not to present an extra challenge to the pose estimator, such as \textit{\enquote{photo, caucasian young male wearing a t-shirt in the city center at daytime sunny day}}.
This template structure can be adapted by the user depending on the target operational domain of the pose estimator.

\paragraph{Variance reduction} 
To reduce variance in our evaluation, the base and attribute images should be as similar as possible, except for the change in the target attribute.
We therefore always use the entire prompt template for generation, \eg, 
we include the same time of day in the prompt even when testing for clothing attributes---this avoids randomly sampling a daytime base image and a nighttime attribute image.
The noise sharing scheme from \cref{sec:problem} further reduces the variation between base and attribute image and thus the evaluation variance. %
The effect can be seen in \cref{fig:gallery}.

\paragraph{Quality control}
To ensure image quality and account for imperfections in the synthesis process, we filter out low-quality generations.
First, we ask the BLIP2~\cite{blip2} visual question answering (VQA) model about the presence of the target attribute, following TIFA \cite{tifa}.
If the answer is negative, the image is discarded.
Second, we use the OpenPose \cite{openpose} 2D pose estimator to check for pose consistency.
If OpenPose's prediction deviates by more than a threshold from the projection of the 3D pose condition, we discard the example.
Since we require paired base and attribute images, we discard the whole pair if either of them is deemed low-quality.
\section{Experiments}
\begin{table*}
\vspace{-0.5em}
\renewcommand{\arraystretch}{1.0}
\setlength{\tabcolsep}{0.74em}
\small
\begin{tabularx}{\linewidth}{lcc|ccccccc}
\multirow[c]{2}{*}{Pose estimator}&\multicolumn{2}{c|}{Base error (mm) $\downarrow$} & \multicolumn{7}{c}{Sensitivity to attributes (percentage of degraded poses, PDP) $\downarrow$}\\
\cmidrule{2-10}
 & MPJPE & PA-MPJPE & {Mean} & {\makecell[l]{Location\\ (outdoor)}}&{\makecell[l]{Location\\ (indoor)}}&{Weather}&{Fairness}&{Clothing}&{Texture}\\ 
\midrule
SPIN~\cite{spin} & 122.50 & 90.27 & 20.70 & 15.79 & 19.65 & 17.43 & 12.93 & 29.13 & 29.26 \\ 
PARE~\cite{pare} &118.81 & 88.98 & 15.27 & 13.34 & 15.84 & 12.97 &\z9.36 & 21.66 &  18.43 \\ 
PyMAF-X~\cite{pymafx}&115.81 & 84.03 & 10.24 & \z9.83 & 12.49 & \z8.51 & \z5.40 & \textbf{14.61}  & \textbf{10.60} \\ 
BEDLAM-CLIFF~\cite{bedlam} &113.14 & 84.22 & 18.65 & 17.30 & 20.77 & 15.47 & 15.61 & 23.36  & 19.42 \\
SMPLer-X-B32~\cite{smplerx} &107.00 & 80.40 & 13.57 & 13.30 & 17.24 & 12.57 & \z7.81 & 17.38 &  13.13 \\ 
SMPLer-X-H32~\cite{smplerx} &104.79 & 76.00 & 12.67 & 14.35 & 18.55 & 10.54 &  \z6.75 & 15.11 & 10.74 \\ 
HMR 2.0~\cite{4dhumans} &102.40 & 75.21 & \textbf{10.19} & \textbf{\z9.31} & \textbf{12.41} & \textbf{\z7.57} & \textbf{\z5.34} & 15.76 & 10.75 \\ 
NLF-S~\cite{nlf} &\z94.04 & 69.91 & 18.21 & 17.35 & 20.15 & 15.86 & 13.86 & 22.94 & 19.08 \\ 
NLF-L~\cite{nlf} &\z \textbf{92.00} & \textbf{67.96} & 18.47 & 17.86 & 20.78 & 17.30 & 13.47 & 22.62 & 18.79 \\
\end{tabularx}%
\vspace{-0.5em}
\caption{\textbf{Evaluating sensitivity of various pose estimators with STAGE}.
We evaluate the percentage of degraded poses (right) and contrast it against the general performance on the base set (left).
We find that appearance attributes such as clothing and texture are the most impactful. In addition, pose estimators are more sensitive to indoor locations compared to outdoor locations. Best performance is highlighted in bold.}
\label{tab:summary}
\end{table*}

\begin{figure*}
\vspace{-1em}
\centering
\small
\setlength{\tabcolsep}{2pt} 
\renewcommand{\arraystretch}{0.8}
\begin{tabular}{cc@{\hskip 0.05in}c@{\hskip 0.13in}cc@{\hskip 0.05in}c@{\hskip 0.13in}cc@{\hskip 0.05in}c}

& \multicolumn{2}{c}{``shirt'' to ``\attrhighlight{floral shirt}'' } &
& \multicolumn{2}{c}{``shirt'' to ``\attrhighlight{parka}'' } & & \multicolumn{2}{c}{``male'' to ``\attrhighlight{female}''} \\

\rotatebox{90}{\hspace{0em} \small BEDLAM CLIFF} & \includegraphics[width=0.145\textwidth]{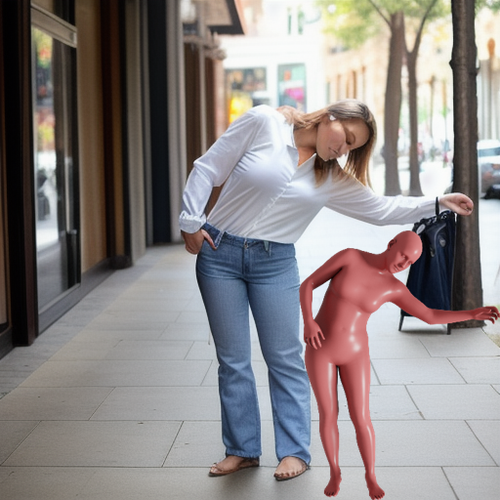} &%
\includegraphics[width=0.145\textwidth]{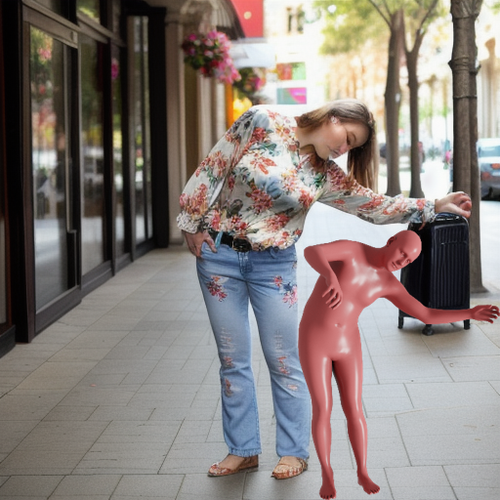} &
\rotatebox{90}{\hspace{0.5em} \small SMPler-X B32} & \includegraphics[width=0.145\textwidth]{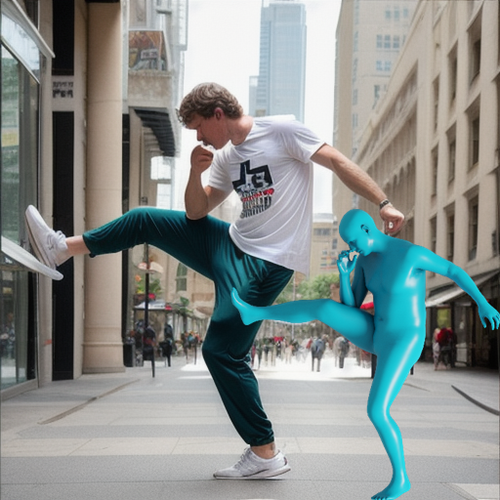} &%
\includegraphics[width=0.145\textwidth]{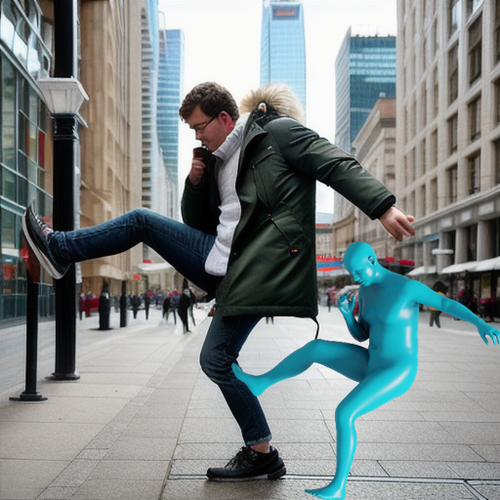} &
\rotatebox{90}{\hspace{0em} \small BEDLAM CLIFF} & \includegraphics[width=0.145\textwidth]{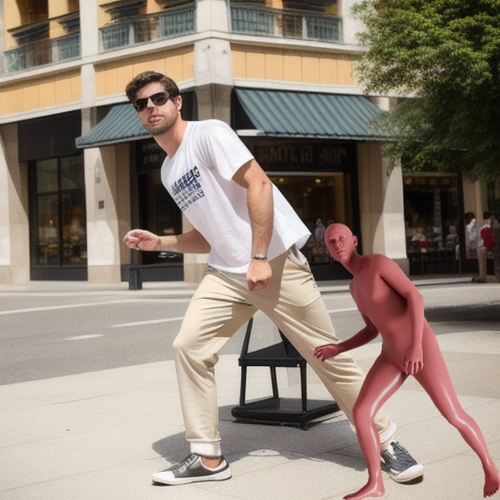} &%
\includegraphics[width=0.145\textwidth]{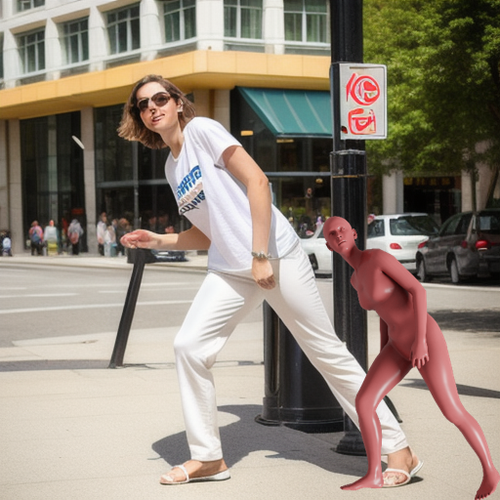}\\[0.7ex]

& \multicolumn{2}{c}{``shirt'' to ``\attrhighlight{checkered shirt}'' } &
& \multicolumn{2}{c}{``shirt'' to ``\attrhighlight{trench coat}'' } & & \multicolumn{2}{c}{``adult'' to ``\attrhighlight{elderly}''} \\

\rotatebox{90}{\hspace{2em} \small PARE} & \includegraphics[width=0.145\textwidth]{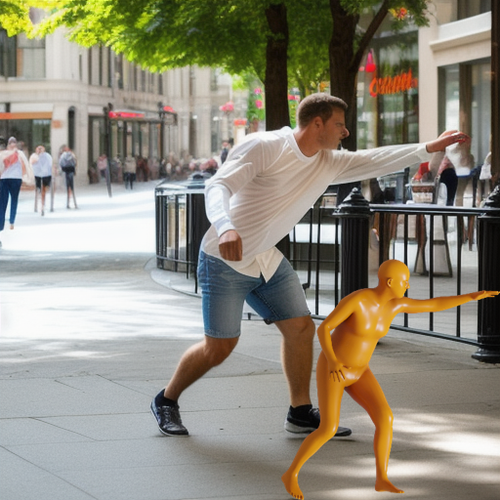} &%
\includegraphics[width=0.145\textwidth]{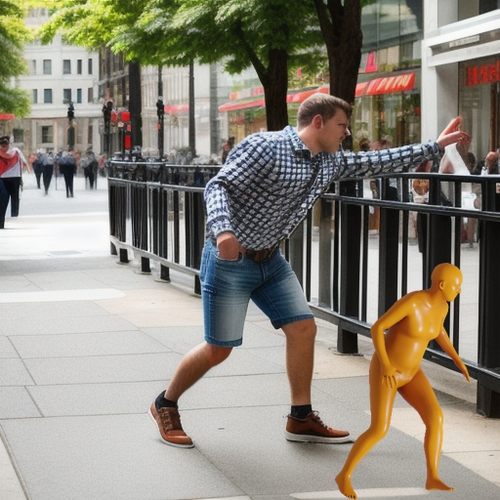} &
\rotatebox{90}{\hspace{1.3em} \small PyMAF-X} & 
\includegraphics[width=0.145\textwidth]{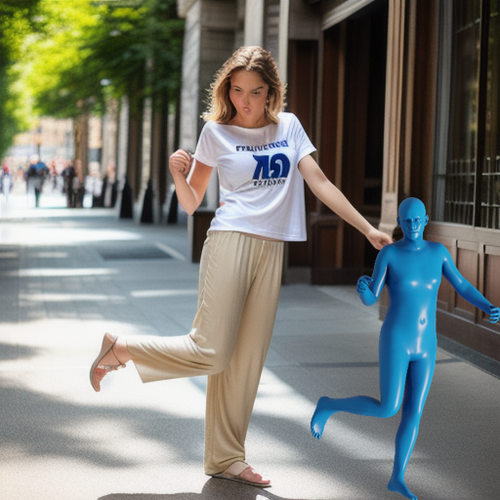} &%
\includegraphics[width=0.145\textwidth]{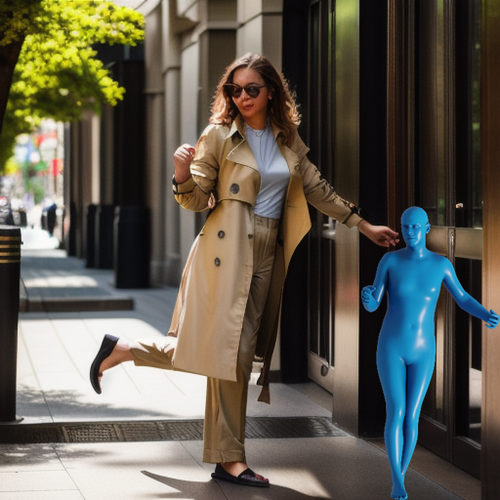} &
\rotatebox{90}{\hspace{0.5em} \small SMPLer-X B32} & 
\includegraphics[width=0.145\textwidth]{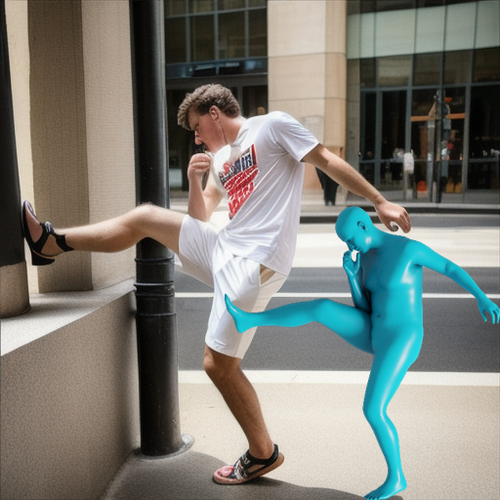} &%
 \includegraphics[width=0.145\textwidth]{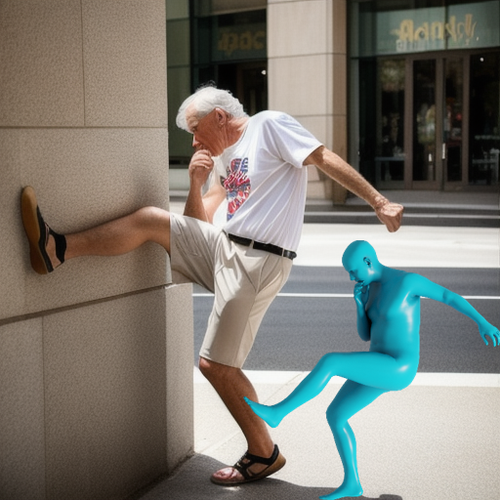}%
\\
& Base image & Attribute image & & Base image & Attribute image & & Base image & Attribute image \\
\end{tabular}
\vspace{-0.5em}
\caption{\textbf{State-of-the-art estimators can break with a simple attribute change}.
For every image pair, we show the predictions of the pose estimator named on the left of the pair.
The base prompt is \emph{``Caucasian adult \{male/female\} wearing a shirt in the city during daytime.''}
}%
\label{fig:exp:texture:fail}
\vspace{-0.9em}
\end{figure*}

\begin{figure}[t]
\begin{centering}
\vspace{-0.4em}
\centering
\small
\setlength{\tabcolsep}{2pt} 
\renewcommand{\arraystretch}{0.8}
\begin{tabular}{ll}
 \rotatebox{90}{\hspace{-2em} \textbf{Clothing}}  & \begin{minipage}{\linewidth} \includegraphics[width=0.94\textwidth]{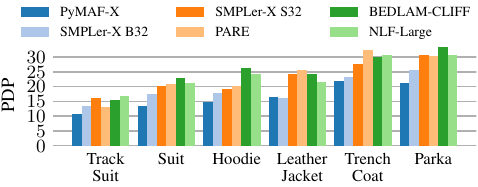} \end{minipage}\\
 \rotatebox{90}{\hspace{-1em} \textbf{Texture}}   & \begin{minipage}{\linewidth} \includegraphics[width=0.94\textwidth]{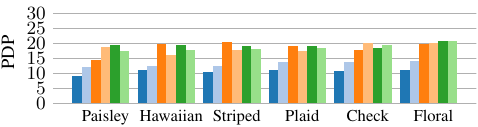} \end{minipage}\\
 \rotatebox{90}{\hspace{-0.6em} \textbf{Fairness}}  & \begin{minipage}{\linewidth} \includegraphics[width=0.94\textwidth]{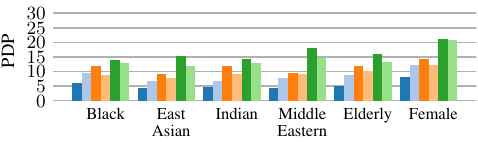} \end{minipage}\\
\end{tabular}
\par\end{centering}
   \vspace{-0.5em}
    \caption{\textbf{Pose estimators are sensitive to individual attribute changes.} For example, clothing texture can have a large impact on performance of pose estimators such as PARE.
    }%
    \label{fig:exp:texture}
    \vspace{-1em}
\end{figure}

In this section, we first use STAGE to evaluate the sensitivity of popular pose estimators towards a selection of attributes.
Then we compare our proposed CN-3DPose generator to GenAI and CG baselines.

\subsection{Examining sensitivity to various attributes}

In the following, we apply STAGE to examine the sensitivity of pose estimators with respect to different open-world attributes.
\cref{fig:gallery} shows a sample of our generated images. Notice how we can synthesize diverse images in a controlled manner. 
We can keep the overall image appearance fixed while changing one specific attribute at a time -- an ideal controlled scenario for evaluation that is unfeasible with purely real data. 
We emphasize that users can choose their own set of attributes for evaluation -- code for generation and evaluation will be made public. 

\paragraph{Experimental setup}
\label{sec:exp:impl}
We initialize our CN-3DPose with weights from CN-Depth~\cite{controlnet} and train it on AGORA~\cite{agora}, HUMBI~\cite{humbi1, humbi2}, SHHQ~\cite{shhq}, COCO~\cite{coco} and scan datasets~\cite{renderpeople, twindom, axzy}.
For evaluation, we pick the popular pose estimators SPIN~\cite{spin}, PARE~\cite{pare}, HMR 2.0~\cite{4dhumans}, PyMAF-X~\cite{pymafx}, SMPLer-X~\cite{smplerx}, NLF~\cite{nlf}, and BEDLAM-CLIFF~\cite{cliff, bedlam}.
Note that BEDLAM-CLIFF achieves state-of-the-art performance by training only on the synthetic AGORA and BEDLAM data~\cite{agora,bedlam}. Here, we examine how effectively this approach generalizes to the open world.

We use 1500 poses for generation.
For experiments on 3DPW, we use farthest point sampling following~\cite{posescript} to obtain diverse poses.
With AMASS, we sample 750 male and 750 female labeled poses from the subset defined by~\cite{posescript}.
The prompts used to create these datasets and further implementation details are provided in the Supp.\@ Mat. 
For evaluation of sensitivity, we use the PDP metric defined in \cref{sec:problem} with $\tau=50\,\mathrm{mm}$.

\paragraph{Summary}
Overall, we observe a significant percentage of degraded poses when testing for a specific attribute as summarized in \cref{tab:summary}.
The attributes that target the body appearance directly (Clothing and Texture) have the most significant effect and can lead to large prediction errors (see \cref{fig:exp:texture:fail}).
Full results are available in the Supp.\@ Mat.

\paragraph{Diverse training data reduces sensitivity}
A natural question is how training data and model size affect the sensitivity of the HPE methods.
Our results indicate that a combination of large-scale data with a large model size is crucial for robustness.
This can be observed on the SMPLer-X variants with different-sized ViT~\cite{vit} backbones.
While they were trained on the same data, only large models use the data effectively, leading to a reduction of PDP across attributes.
This is most pronounced for attributes of fairness, clothing, and clothing texture attributes, \ie, attributes that affect the appearance of the person.
HMR 2.0 is a piece of further evidence for that, as it was trained on a large-scale dataset with a ViT-H backbone.
Conversely, SPIN and PARE were trained with the least amount of data and are the most sensitive estimators (largest PDP).
However, diversity alone is not enough.
BEDLAM-CLIFF (SOTA on real benchmarks) was trained on purely synthetic data, designed to be highly diverse, but is very sensitive to attribute changes across the board.
This further evidences that our STAGE method sheds new light on the performance of SOTA methods in the open world, which can not be measured in existing benchmarks. 

\paragraph{Architectural choices matter}
While large-scale datasets lead to lower PDP, the performance of PyMAF-X indicates that smart architectural choices can be an efficient way to achieve robustness as well.
PyMAF-X was trained on roughly the same data as PARE but is less sensitive to attribute changes.
We attribute this to the iterative feedback loop to align the prediction to the input image in PyMAF-X.
While the general performance is only about 3~mm better than PARE, it results in a significantly more robust method, which was impossible to verify before STAGE.

\paragraph{Location and weather}
It is commonly accepted in the HPE community that outdoor scenes are more challenging than indoor scenes~\cite{eft}. However, our results show that SOTA methods degrade more when the attribute is indoor; see \cref{tab:summary}. We reckon that indoor benchmarks are typically captured in lab settings with very simple backgrounds, whereas our generated images have more realistic backgrounds, including potential clutter and occlusions. This hints that indoor HPE in the open world is not necessarily easier than outdoors.
Perhaps unsurprisingly, bad weather conditions, in particular snow, result in the most degradation of poses.   

\paragraph{Clothing significantly affects pose estimation}
While intuitive, for the first time, we empirically verify that clothing strongly affects HPE performance. 
\cref{fig:exp:texture} shows the sensitivity to clothing items and different texture patterns.
We observe that large body covering items such as coats and long jackets are most impactful, leading to a degradation of up to 30\% of the predictions and over 15\% across all estimators.
In addition, a simple change such as the pattern of a shirt can lead to up to 20\% of degraded predictions.

\paragraph{Fairness}
\cref{fig:exp:texture} also shows the sensitivity to protected attributes -- attributes that should not be used to make decisions. 
Overall, current pose estimators are less sensitive towards protected attributes (such as ethnicity) compared to clothing and location. 
This might be due to the fact that these attributes usually only affect a small part of the body, making them less important overall.
We note, however, that gender does seem to affect predictions more (see also \cref{fig:exp:texture:fail}) than other attributes -- which indicates that to make these methods fair, more work is needed.
One can also observe that BEDLAM-CLIFF is quite sensitive to every attribute despite being designed to cover a range of skin tones.
This indicates that CG-generated data might not be sufficient to achieve good open-world performance.

\begin{table}[t]
    \vspace{-1em}
    \small
    \noindent
\setlength{\tabcolsep}{1.5pt}
\renewcommand{\arraystretch}{1.0}
    \begin{tabular}{@{}l|ccccc@{}}
    \multirow{2}{*}{Image generator}  & \multicolumn{2}{c}{Pose gap $\downarrow$}  & \multirow{2}{*}{FID$\downarrow$}  & 3D pose & \multirow{2}{*}{Diversity} \\
    & \scriptsize w/ backgr. & \scriptsize w/o backgr. &  & coherence & \\
        \midrule
        CN-Pose            &  32.83 & 26.49  & 58.8 & \xmark & \xmark \\
        CN-Multi           & \z9.19 & \z8.10 & 60.8  & \cmark & \xmark \\
        CN-3DPose  (ours)  & 13.11  & \z9.65 & 39.7 & \cmark & \cmark \\
    \end{tabular}
    \caption{
    \textbf{GenAI Baselines.}
    Our CN-3DPose generates more realistic images (FID) with comparable 3D pose alignment (pose gap is the difference in PA-MPJPE between real and generated images).}
    \label{tab:baseline}
    \vspace{-0.7em}
\end{table}

\begin{figure}
\vspace{-0.8em}
\setlength{\tabcolsep}{0pt} 
\renewcommand{\arraystretch}{0.8}
\centering
\scriptsize
\begin{tabular}{@{}c@{\hskip 0pt }c@{\hskip 1pt}c@{\hskip 2pt}c@{\hskip 2pt}c}
&&\multicolumn{3}{c}{Photo, adult caucasian male wearing ...} \\
&
&
\pbox{0.22\linewidth}{jacket in city} &%
\pbox{0.22\linewidth}{\centering coat in snowy city} &%
\pbox{0.22\linewidth}{jacket in restaurant}\\
\multirow[l]{2}{*}[-2.0em]{%
    \includegraphics[width=0.23\linewidth]{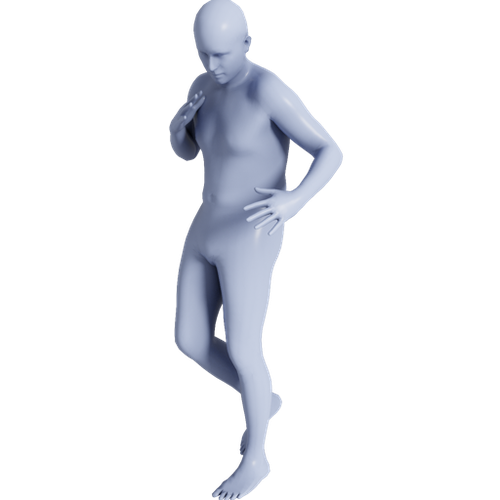}
} &%
\rotatebox{90}{\hskip 2.4em CN-Pose} &%
\includegraphics[width=0.24\linewidth]{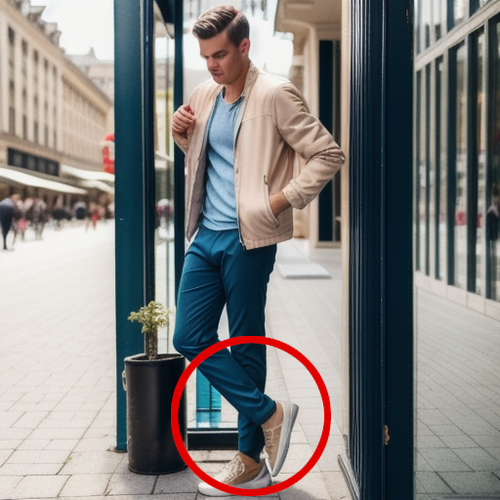} &%
\includegraphics[width=0.24\linewidth]{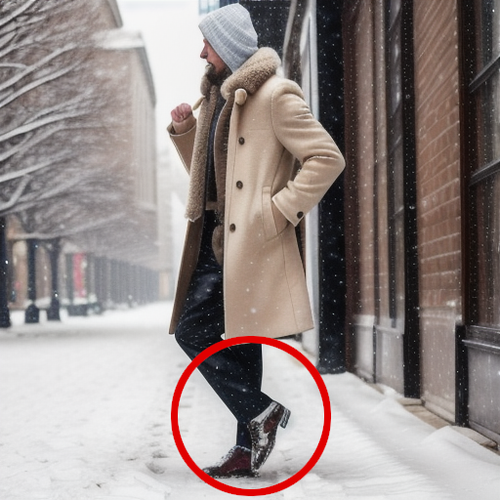} &%
\includegraphics[width=0.24\linewidth]{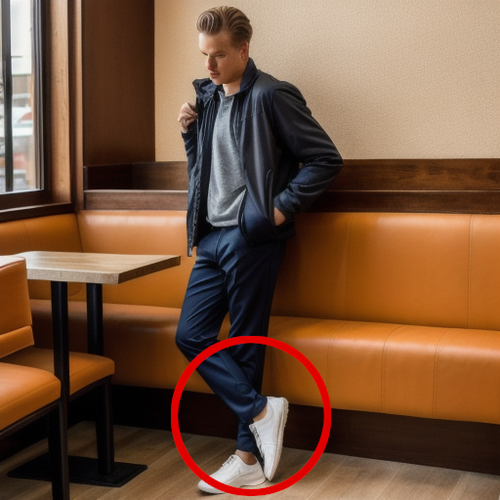} %
\\[3pt]
\multirow[c]{1}{*}[-1.2em]{%
    \pbox{0.1\linewidth}{\centering GT pose}
} 
&
\rotatebox{90}{\hskip 2.3em CN-Multi}  &%
\includegraphics[width=0.24\linewidth]{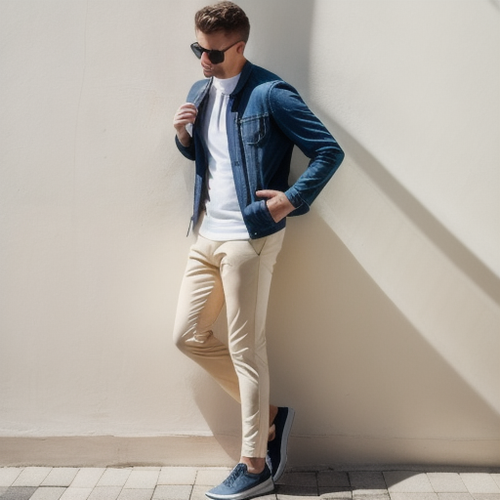} &%
\includegraphics[width=0.24\linewidth]{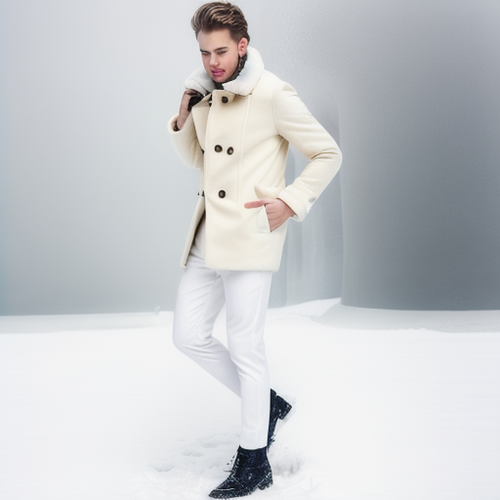} &%
\includegraphics[width=0.24\linewidth]{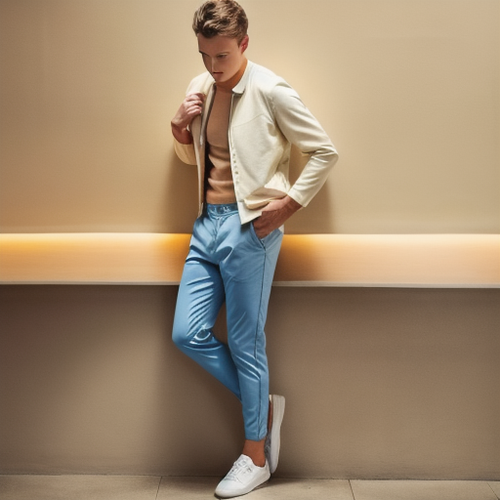} %
\\[3pt]
&
\rotatebox{90}{\hskip -1pt CN-3DPose (Ours)} &%
\includegraphics[width=0.24\linewidth]{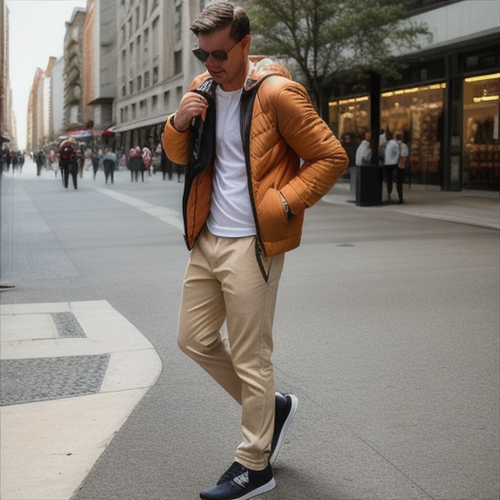} &%
\includegraphics[width=0.24\linewidth]{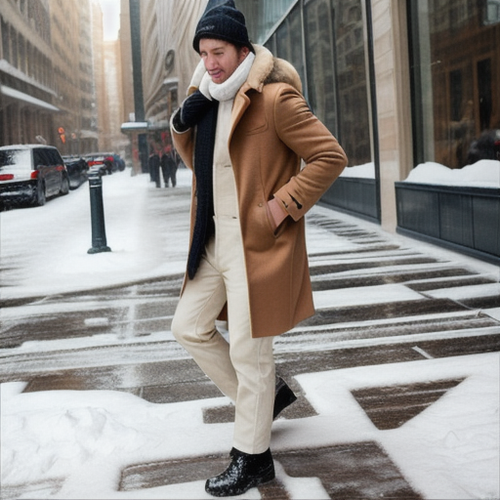} &%
\includegraphics[width=0.24\linewidth]{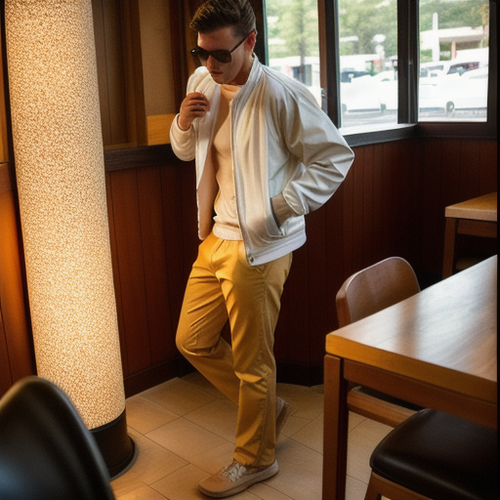} %
\\[1pt]
\end{tabular}
\vspace{-0.5em}
\caption{
\textbf{Our method CN-3DPose generates images that are both diverse and pose-aligned.}
In contrast, CN-Pose consistently generates the feet in the wrong depth order and CN-Multi generates flat backgrounds and limited clothing variation.
}%
\label{fig:oursvs:imgs}
\vspace{-2em}
\end{figure}

\subsection{Comparison against GenAI and CG baselines}
\label{sec:exp:quality}

\paragraph{GenAI models}
We first compare our CN-3DPose to the GenAI baselines CN-Pose and CN-Multi from \cref{sec:model} regarding pose coherence and image diversity.
We use each of these image generators to construct a synthetic replica of the 3DPW benchmark~\cite{3dpw}, which covers diverse real-world scenes such as parks, city centers, and restaurants.
We aim for a similar attribute distribution, \eg, same clothing, gender, and location.
This way, the dominant distribution shift comes from the model used to generate the images. %
We compare: 1) visual quality and diversity through the FID~\cite{fid} score and 2) pose coherence, by running a SOTA pose estimator~\cite{metrabs_acae} on all images and computing the difference between PA-MPJPE achieved on generated images \vs on real data, which we refer to as \emph{pose gap}.
This measures whether the pose in the generated image is recognizable to the estimator to a similar degree as in real images.
To extract attributes from a real 3DPW image, we use a VQA model~\cite{blip2}. See \suppmat{} for details.

We present the results in \cref{tab:baseline}. 
Our CN-3DPose is significantly better than CN-Pose in pose coherence, yielding a pose gap of 13.11 mm \vs 32.83. %
This reinforces our observation that 3D consistency requires 3D conditioning. 
For example, as seen in \cref{fig:oursvs:imgs}, CN-Pose suffers from swaps in the depth order of the feet.
While CN-Multi generates images with good 3D consistency, it is biased towards flat backgrounds and body-tight clothing (see \cref{fig:oursvs:imgs} and \suppmat{}), leading to a worse FID score compared to our method (60.8 \vs 39.7). 
Note that computing the pose gap on full images favors generators that produce simple images, making the pose estimator's task easier.
Hence, a lower pose gap does not always imply better images.
We therefore also compute the pose gap using just the foreground. 
Although the pose gap of the simpler images of CN-Multi barely changed (9.19 $\Rightarrow$ 8.10), simplifying the backgrounds of CN-3DPose significantly decreases the pose gap (13.11 $\Rightarrow$ 9.65). 
We hypothesize that this bias in the metric explains CN-Multi's slight advantage in pose gap---the estimator is not robust to the complex clothing of our CN-3DPose.
We therefore conclude CN-3DPose is the best choice to generate images for controlled auditing of 3D HPE methods.

\begin{table}[t]
\vspace{-1em}
{\small
{\setlength{\tabcolsep}{2pt}
\renewcommand{\arraystretch}{1.0}
\begin{tabularx}{\linewidth}{@{}lcc|cc|cc@{}}
\multirow{2}{*}{Method} & \multicolumn{2}{c|}{Training data} &  \multicolumn{2}{c|}{3DPW-test} &  \multicolumn{2}{c}{EMDB-test} \\
       &     \makecell{CG}    & STAGE         &  MPJPE &  PA-MPJPE &  MPJPE &  PA-MPJPE \\
\midrule
  \multirow{3}{*}{PARE} &       \cmark             &               &  76.24 &     49.07 &  98.55 &     60.11 \\
  &                          &        \cmark &  75.37 &     48.52 &  97.32 &     61.76 \\
  &       \cmark             &        \cmark &  72.93 &     46.84 &  92.71 &     57.65 \\
  \midrule
   \multirow{3}{*}{HMR} &       \cmark             &               &  75.07 &     48.08 &       96.61 &          59.08 \\
  &                          &        \cmark &  76.23 &     48.72 &       96.66 &          62.62 \\
  &       \cmark             &        \cmark &  72.75 &     46.80 &       91.56 &          58.02 \\
\end{tabularx}
}
}
\vspace{-0.5em}
\caption{\textbf{Training on STAGE data rivals training on time-consuming CG-data designed by human experts.}}
\label{tab:quality:train}
\vspace{-1em}
\end{table}

\paragraph{STAGE vs. BEDLAM}
In order to compare against CG baselines in this setting, we would need to create a simulated replica of the 3DPW dataset. 
The cost of skilled labor to create such high-quality CG renderings is too high to be feasible.
We thus propose to compare the data quality by evaluating the generalization performance of pose estimators trained on it. 
For a CG baseline, we chose the high-quality BEDLAM~\cite{bedlam} dataset.
BEDLAM enables methods to achieve SOTA performance on real-world benchmarks by using hundreds of assets and scenes laboriously crafted by artists to mimic real life.
To show that the STAGE data can enable similar performance, without laborious manual effort, we use the body poses of BEDLAM while relying on CN-3DPose for diversity and realism.
We train HMR~\cite{hmr} and PARE~\cite{pare} on BEDLAM's CG images, on STAGE, and a combination of both. We then evaluate on the common 3DPW \cite{3dpw} and the recent EMDB \cite{emdb} benchmarks using MPJPE and PA-MPJPE. Further details on training configuration are in \suppmat{}

\begin{figure}%
\small
\setlength{\tabcolsep}{0.1em}
\begin{tabular}{@{}c@{\hskip 2pt}cccc@{}}
\rotatebox{90}{\hskip 0.8em BEDLAM } &
\includegraphics[width=0.23\linewidth]{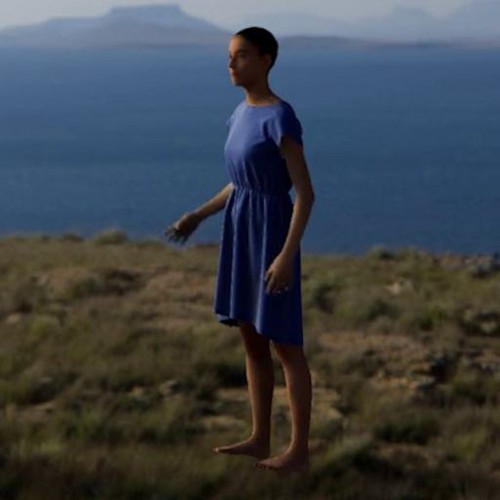} &%
\includegraphics[width=0.23\linewidth]{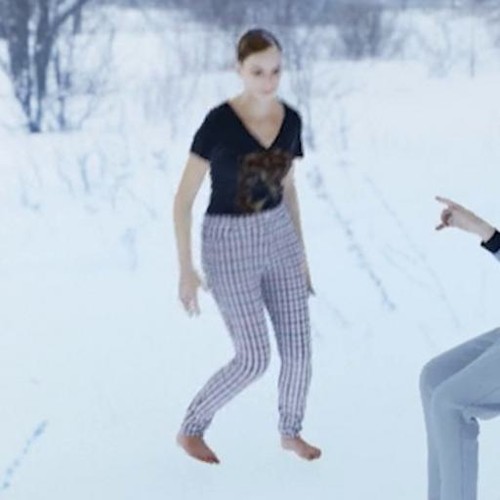} &%
\includegraphics[width=0.23\linewidth]{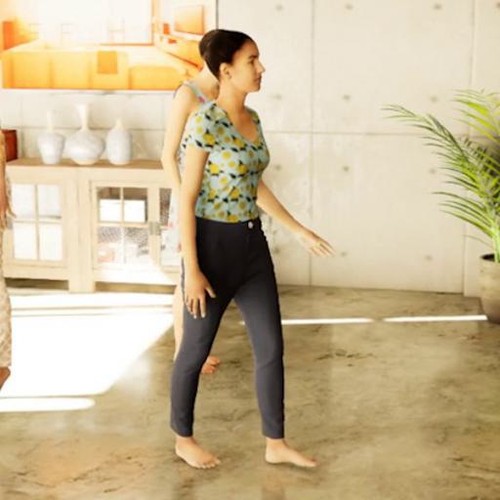} &%
\includegraphics[width=0.23\linewidth]{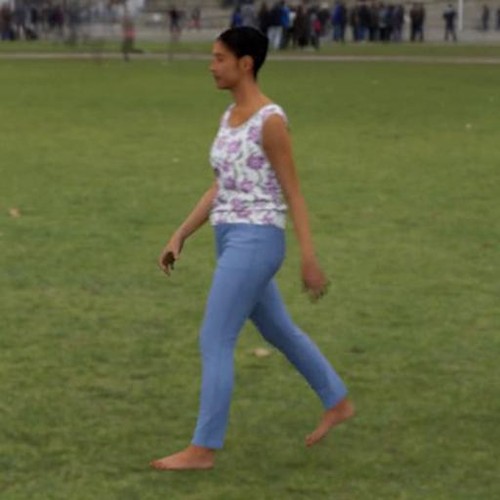} \\
\rotatebox{90}{\makecell{\makebox[0pt][l]{STAGE (ours)}}} &
\includegraphics[width=0.23\linewidth]{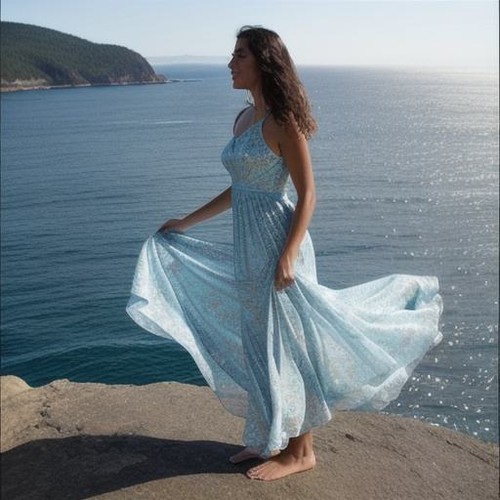} &%
\includegraphics[width=0.23\linewidth]{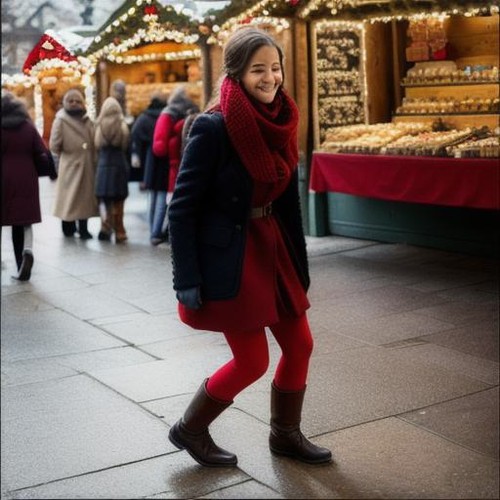} &%
\includegraphics[width=0.23\linewidth]{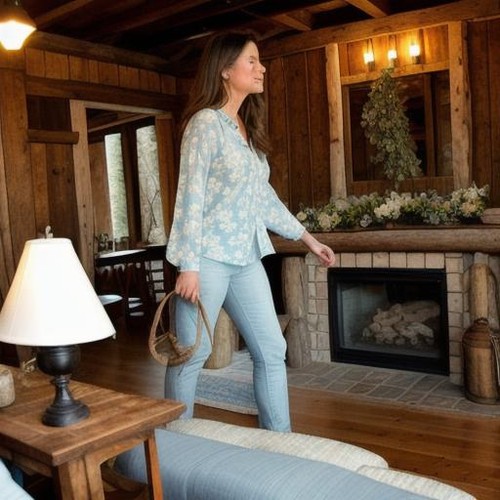} &%
\includegraphics[width=0.23\linewidth]{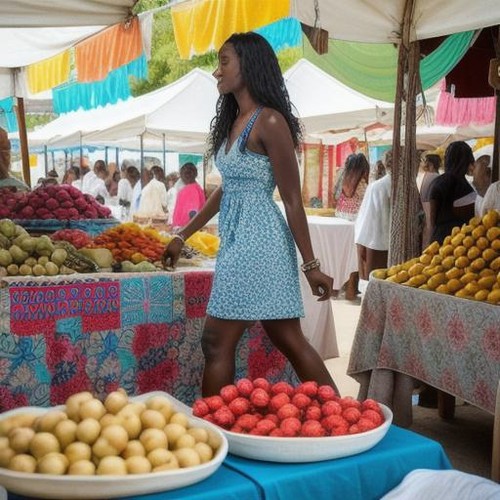} \\
\end{tabular}%
\vspace{-0.5em}
\caption{
\textbf{Comparison of STAGE and BEDLAM\@.} %
STAGE images depict more realistic clothing and locations. Note especially the physically plausible wind effects and the complex scene composition; both of which would be costly to create in a simulation. 
}%
\label{fig:bedlam:vs:ours}%
\vspace{-1.0em}
\end{figure}

The results are presented in \cref{tab:quality:train}.
Overall, models trained on STAGE data can match the performance of models trained using BEDLAM.
This shows that our data quality is good enough to achieve the state of the art while our text-to-image model is significantly cheaper and easier to scale than traditional CG pipelines for scene and asset creation.
Finally, training on the combination of STAGE and BEDLAM results in improved performance across all models. This indicates that the GenAI data complements the CG data providing extra diversity, while CG data gives extra precision over image and body pose alignment. 

\paragraph{Limitations} In this work, we focus on evaluating single-person, single-image 3D human pose estimation. 
We do not consider interpersonal or human--object interactions, or video. 
These would be interesting future extensions. 
Furthermore, current image generators are not perfect, but our modular framework allows for easy integration of more advanced future versions.
\section{Conclusion}
3D HPE has been an active research area for over 30 years and has seen a lot of progress in terms of generalization and good benchmarks. However, are these methods robust to attributes such as clothing, weather, or gender? 
For the first time, we can answer these questions empirically, shedding new light on HPE methods with STAGE, a toolkit to generate custom benchmarks for 3D HPE at low cost.
We build upon text-to-image models and adapt their capabilities to generate diverse, realistic images with 3D human pose and attribute control. 
With this model, we create synthetic benchmarks and audit
current SOTA pose estimators against attributes such as clothing, texture, location, weather, and fairness.
We make several interesting findings with our method: most estimators are sensitive to clothing (intuitive but never verified empirically) and texture. To a lesser degree, most methods are also affected by certain protected attributes such as ethnicity, gender, or age. Overall, the methods are sensitive to multiple naturally occurring attributes; this is concerning and requires further investigation to make HPE methods fair and safe. 
We will release code and data to allow researchers and practitioners to evaluate their 3D HPE methods against their desired attributes, allowing a much deeper understanding of how they will perform in the target operational domain.

\paragraph{Acknowledgments:}
We thank Riccardo Marin for proofreading and the whole RVH team for the support.
Nikita Kister was supported by Bosch Industry on Campus Lab at the University of Tübingen.
Nikita Kister thanks the European Laboratory for Learning and Intelligent Systems (ELLIS) PhD program for support.
István Sárándi and Gerard Pons-Moll were supported by the German Federal Ministry of Education and Research (BMBF): Tübingen AI Center, FKZ: 01IS18039A, by the Deutsche Forschungsgemeinschaft (DFG, German Research Foundation) -- 409792180 (Emmy Noether Programme, project: Real Virtual Humans).
GPM is a member of the Machine Learning Cluster of Excellence, EXC number 2064/1 -- Project number 390727645 and is supported by the Carl Zeiss Foundation.

{
    \small
    \bibliographystyle{ieeenat_fullname}
    \bibliography{abbrev_short,pose,main,people,evaluation}
}

\clearpage
\renewcommand{\thepage}{S\arabic{page}}

\setcounter{section}{0}
\setcounter{figure}{0}
\setcounter{table}{0}
\setcounter{equation}{0}
\setcounter{page}{1}
\maketitlesupplementary

\renewcommand{\thesection}{\Alph{section}}
\renewcommand{\thefigure}{S\arabic{figure}}
\renewcommand{\thetable}{S\arabic{table}}
\renewcommand{\theequation}{\alph{equation}}
\newcommand{\promptstring}[1]{\textit{\enquote{#1}}}

\section{Implementation details}
\subsection{Training CN-3DPose}
The CN-3DPose architecture is based on ControlNet~\cite{controlnet}, where we adapt the input layer to take 9 channels, 3 for each of the 3 input conditions (dense SMPL semantic encoding, 2D skeleton drawing, depth).
We initialize the weights from a pre-trained CN-Depth~\cite{controlnet} and train 
on AGORA~\cite{agora}, HUMBI~\cite{humbi1}, SHHQ~\cite{shhq}, COCO~\cite{coco} and a set of human scan renderings~\cite{twindom, renderpeople, axzy}.

We use image crops of size 512$\times$512~px.
We remove images where the person is significantly occluded or truncated.
For COCO, we require at least 10 visible keypoints; for AGORA we require 20\% of SMPL vertices to project within the image boundary.
When the person is near the edge of the image, we use zero padding to ensure the person is centered in the crop.
We mask these padding regions in the loss computation, so the model does not learn to generate them at test time.
Since we focus on single-person scenarios in this work, we also mask out other people's regions from the loss, using the masks provided in the datasets.

\paragraph{Caption generation}
CN-3DPose is a pose-conditioned text-to-image model.
To train it, we need image captions in the training data.
We generate these using BLIP2~\cite{blip2}.
For each image, we sample five captions and filter them using CLIP~\cite{clip} to ensure image-text alignment.
For COCO, we also use the captions provided in the dataset annotations.
During training, we randomly pick a caption from this filtered set for each image.

We use some renderings in our training data (AGORA and the scan renders), but would like to avoid reproducing their distinct CG look at test time.
To this end, we add the word \promptstring{Rendering} to the captions of rendering-based training images, and use negative prompting with this word at inference time.

\paragraph{Training details}
For all hyperparameters, we follow \cite{controlnet}.
For batch construction we sample examples from each dataset with probability proportional to the size of the dataset.
We train on 4 NVIDIA A100 (40GB) for nine days.

\subsection{2D pose estimation-based filtering}
To identify low-quality generations we use OpenPose~\cite{openpose} to predict 2D keypoints and compare them to the ground truth.
We convert the SMPL-based ground-truth pose to the OpenPose/COCO format using a joint regressor on the SMPL vertices. 
The image is discarded if the projection error exceeds 50~px for at least one wrist, ankle, shoulder, elbow, or knee keypoint, or OpenPose detects multiple people in the image.

\subsection{Replica of 3DPW}
\label{suppl:sec:synth:replica}
To validate our image quality, we create a synthetic replica of a subset of 3DPW.
We sample 1500 examples based on farthest point sampling of the poses to obtain a diverse subset.
We use the same camera angles as in 3DPW.
To create the text prompts, we use the prompt template \promptstring{Photo, caucasian \{gender\} wearing \{clothing\} in \{location\} during a sunny day at daytime}.
We fill in the gender from the annotations, and use BLIP2~\cite{blip2} to describe clothing and location, by asking 
\promptstring{What is the person in the foreground wearing?} and \promptstring{Where is the person located?}.

We use 2D pose-based and VQA-based quality checks, and regenerate low-quality samples.
However, if we exceed 13 attempts for a given pose, we discard that pose.
The number of resulting valid poses is 1021 for CN-Pose, 1367 for CN-Multi and 1372 for CN-3DPose.
For fair comparison between them, we only use the common subset of these, containing 981 examples.

\subsection{Data generation for attribute experiments}
\paragraph{Camera settings}
For the attribute robustness experiments, we use 1500 AMASS poses as described in the main paper.
Since AMASS does not contain camera information, we construct the camera parameters as follows.
The camera faces the root joint frontally, such that the wrists, ankles, shoulders, and elbows are visible and the distance is such that the pose fits tightly within the frame.

\paragraph{Base prompt}
Generally we use the base prompt \promptstring{Photo, caucasian \{gender\} wearing t-shirt and pants in city center at daytime sunny weather}, where gender is filled in based on the pose that is used for generation.
However, for some specific experiments, we need to make adjustments. 
For the indoor location experiment, the base prompt has \promptstring{hallway} instead of \promptstring{city center} and in the outdoor location experiment we use \promptstring{village} in the base prompt to evaluate \promptstring{city center} itself.
For gender, we use \promptstring{male} in the base set and \promptstring{female} for the attribute set.
For body shape, we use \promptstring{adult with average BMI} for base and  \promptstring{adult with low/high BMI} for the two attribute datasets.

\paragraph{Quality filtering}
We again apply 2D pose and VQA-based quality checks and filtering.
For fair comparison between attributes of a category, we use only poses that resulted in valid images for all attributes of that category.

\begin{figure*}[t]
\begin{centering}
\begin{tabular}{cc}
\includegraphics{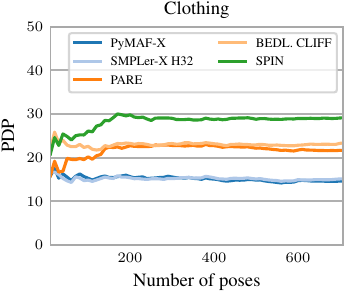} &%
\includegraphics{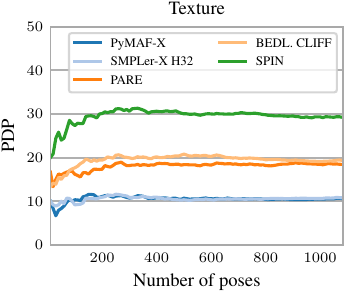} %
\\
\includegraphics{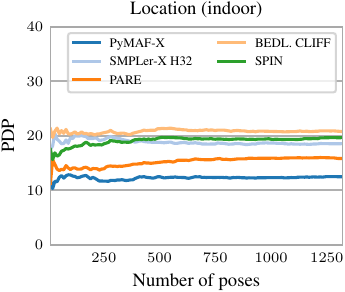} &%
\includegraphics{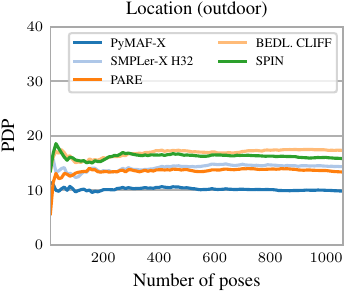} %
\\
\includegraphics{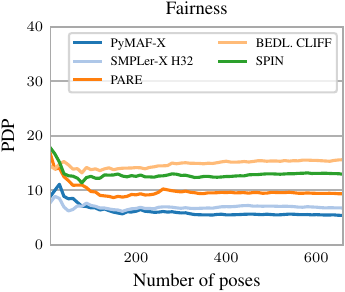} &%
\includegraphics{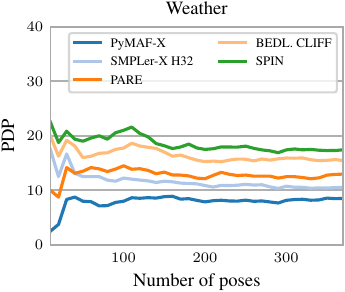} %
\\
\end{tabular}
\par\end{centering}
\caption{
\textbf{Our number of samples is sufficient.}
We plot how our results per category change based on the number of valid poses used to compute them. Overall the PDP remains stable after the first couple of hundred of poses.
}%
\label{fig:supp:numsamples}
\end{figure*}

The number of resulting poses is listed in \cref{tab:supp:numsamples}.
To make sure that this number of poses is enough for a reliable calculation of the percentage of degraded poses (PDP), we investigate how the PDP changes with different numbers of samples. 
As seen in \cref{fig:supp:numsamples}, the PDP is stable after the first couple of hundred poses, confirming that the number of samples is sufficient.
\begin{table}
\centering
\small
\setlength{\tabcolsep}{2pt}
\begin{tabularx}{\linewidth}{@{}ccccccc@{}}
    \toprule
     & \makecell{Location \\(outdoor)} & \makecell{Location \\(indoor)} & Fairness & Clothing & Weather & Texture \\
     \midrule
    \makecell{\# poses } & 1062 & 1320 & 660 & 708 & 370 & 1085 \\
    \bottomrule\\
\end{tabularx}
\vspace{-1em}
\caption{\textbf{Number of poses used in the robustness experiments}. 
We use several hundred to more than a thousand examples to compute PDP for each attribute.
The numbers differ between categories because we filter out low-quality generations. In each category, we only keep those poses for which generation was successful for every attribute of that category. This ensures that PDP can be compared reliably among the different attributes of a category.
}
\label{tab:supp:numsamples}
\end{table}

\section{BEDLAM vs. STAGE}
\paragraph{Training data generation with CN-3DPose}
Since BEDLAM is a video dataset, the poses for consecutive frames can be very similar.
To reduce redundancy in the data, we therefore pick poses only if at least one joint has moved by at least 10 cm.
We further discard highly occluded instances that cannot be detected by YOLOv4.
This filtering results in a set of 355k annotated image crops.
For each of these examples, we then generate a corresponding image using our CN-3DPose model using the pose annotation for conditioning.
For the text prompt, we populate our prompt template from \cref{suppl:sec:synth:replica} with random combinations of the attributes used in our robustness experiments.
Similar to the robustness experiments, here we also perform quality checks based on OpenPose's 2D estimation, and discard generations where OpenPose has high error.
The result is a dataset with the same body pose and camera distribution as BEDLAM, but with more diverse and more realistic images.

\paragraph{Pose estimator training details}
We use MMHuman3D~\cite{mmhuman3d} to train HMR and PARE---both of them with the HRNet-W48 \cite{hrnet} backbone pretrained on COCO keypoints. We train for 50 epochs with a batch size of 128 for HMR and 64 for PARE.
The learning rate starts at $3.3\times 10^{-5}$ for HMR and at $1.6\times 10^{-5}$ for PARE, then decays exponentially with a rate of $0.9729$ for the first 40 epochs and $0.896$ for the last 10 epochs.
We use a single NVIDIA A100 (40GB) GPU.

\section{Full results}
\begin{figure*}[t]
    \centering
    \includegraphics[width=0.9\textwidth]{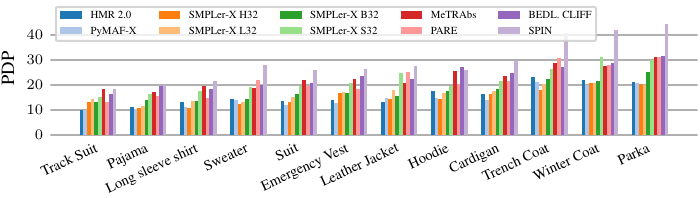}
    \caption{\textbf{Clothing impacts performance.} Clothing has the largest impact on performance. Especially items that cover most of the body, such as coats, can impact the performance in a negative way.}
    \label{fig:supp:clothing}
\end{figure*}

\begin{figure*}[t]
    \centering
    \includegraphics[width=0.90\textwidth]{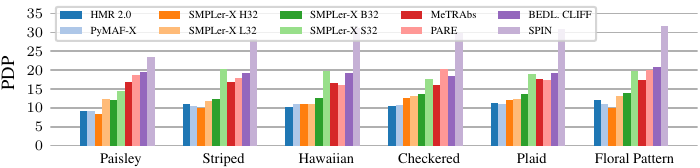}
    \caption{
    \textbf{Pose estimators are susceptible to texture changes.} All textures lead to about the same PDP, indicating that a texture change influences performance regardless of what that texture is. SPIN is particularly sensitive as up to 30 percent of the poses are degraded.
    }%
    \label{fig:supp:texture}
\end{figure*}

\begin{figure*}[t]
    \centering
    \includegraphics[width=0.90\textwidth]{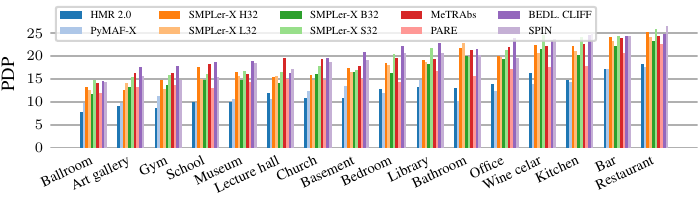}
    \caption{\textbf{Influence of indoor locations.} The continual increase of the PDP indicates that some locations are more challenging than others. Especially, ``Restaurant'', ``Bar'' and ``Wine cellar'' have the most impact on performance.
    }%
    \label{fig:supp:location:in}
\end{figure*}

\begin{figure*}[t]
    \centering
    \includegraphics[width=0.90\textwidth]{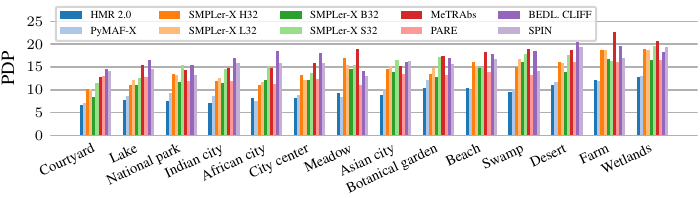}
    \caption{\textbf{Influence of outdoor locations}. Most outdoor locations have a similar impact on performance, ``Swamp and ``Wetlands'' pose the most challenges to pose estimators.
    }%
    \label{fig:supp:location:out}
\end{figure*}

\begin{figure*}[t]
    \centering
    \includegraphics[width=0.90\textwidth]{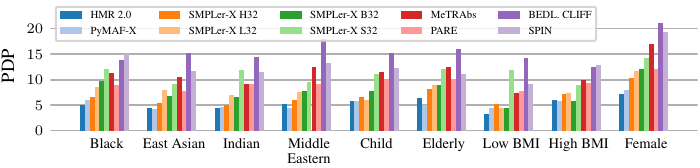}
    \caption{
    \textbf{Fairness analysis.} We consider multiple attributes related to fairness in computer vision. Pose estimators are robust against protected attributes. However, gender and age.
    }%
    \label{fig:supp:fairness}
\end{figure*}

\begin{figure*}[t]
    \centering
    \includegraphics[width=0.90\textwidth]{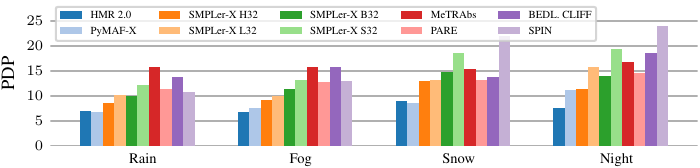}
    \caption{\textbf{Influence of adverse conditions}. Adverse conditions such as snow and night influence the performance. Pose estimators seem less sensitive to fog and rain.
    }%
    \label{fig:supp:weather}
\end{figure*}

\begin{figure*}%
\begin{centering}
\begin{tabular}{@{}ccccc}
\includegraphics[width=0.18\linewidth]{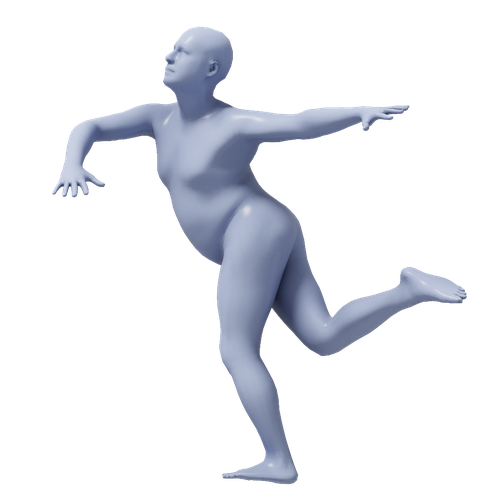} &%
\includegraphics[width=0.18\linewidth]{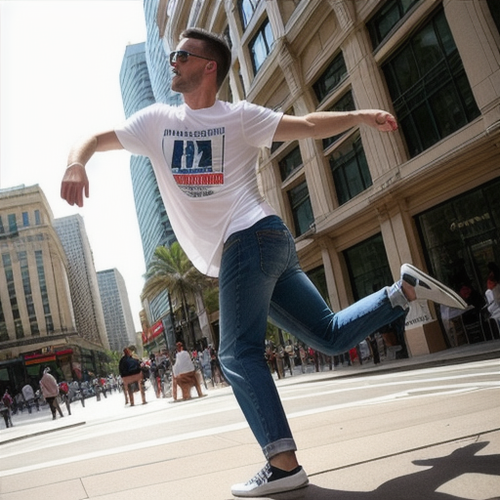} &%
\includegraphics[width=0.18\linewidth]{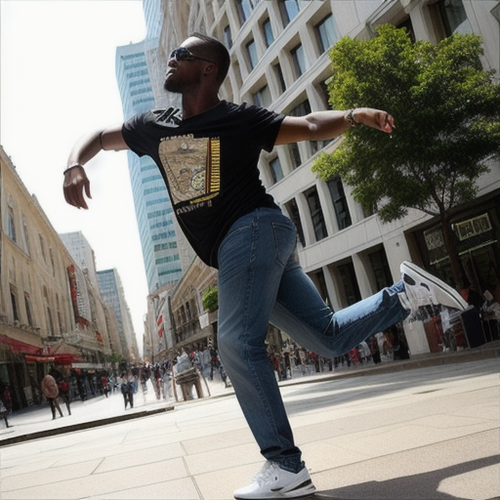} &%
\includegraphics[width=0.18\linewidth]{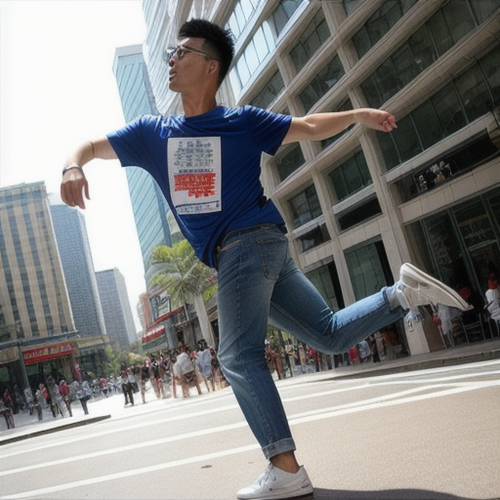} &%
\includegraphics[width=0.18\linewidth]{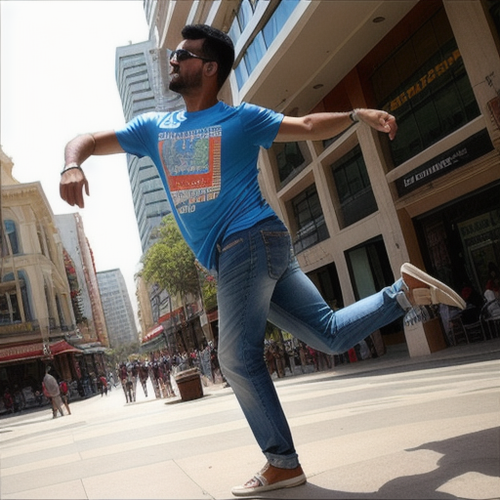} %
\\[-3pt]
& Caucasian & Black & East Asian & Indian  \\[12pt]
\includegraphics[width=0.18\linewidth]{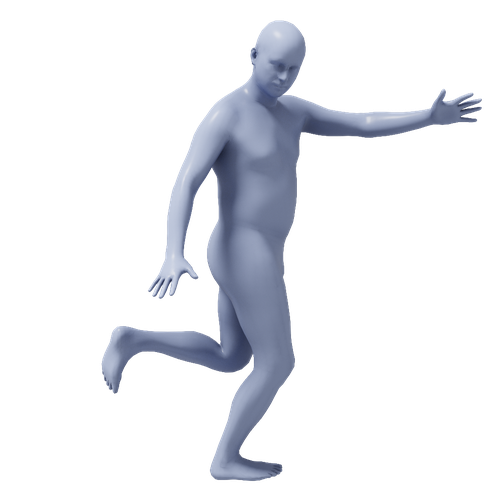} &%
\includegraphics[width=0.18\linewidth]{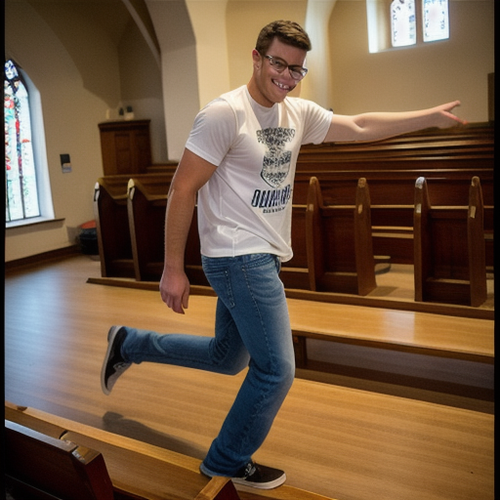} &%
\includegraphics[width=0.18\linewidth]{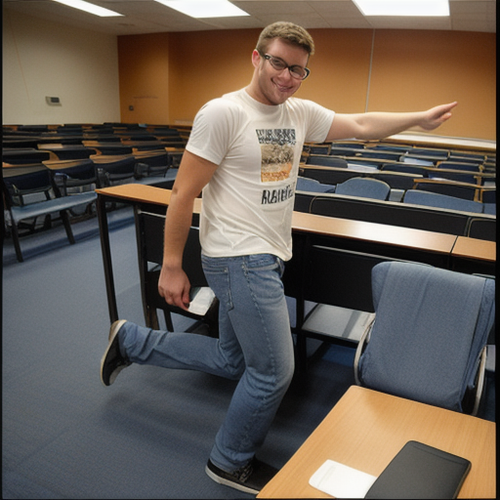} &%
\includegraphics[width=0.18\linewidth]{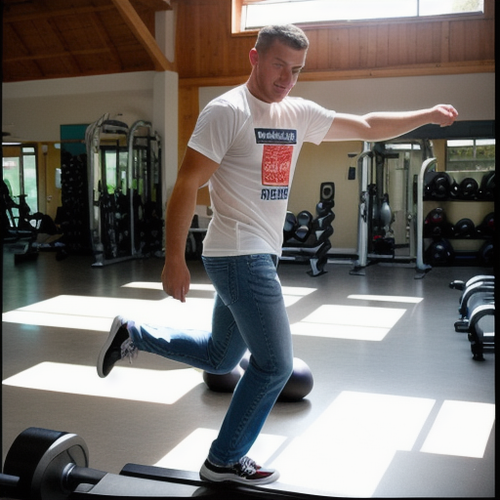} &%
\includegraphics[width=0.18\linewidth]{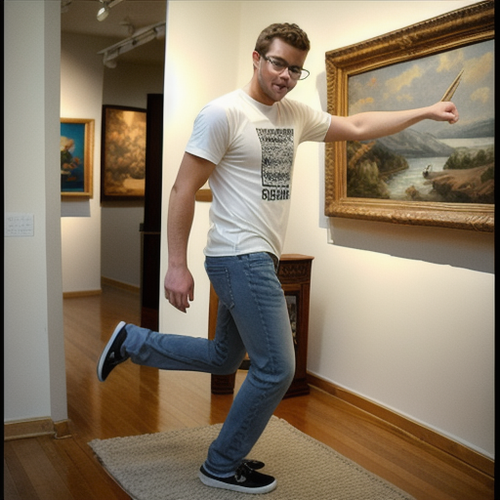} %
\\[-3pt]
& Church & Lecture hall & Gym & Art gallery  \\[12pt]
\includegraphics[width=0.18\linewidth]{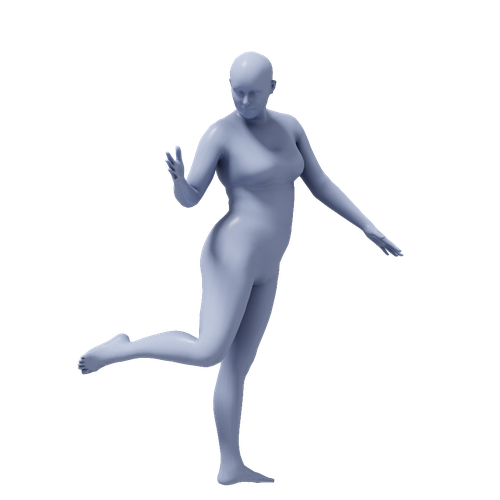} &%
\includegraphics[width=0.18\linewidth]{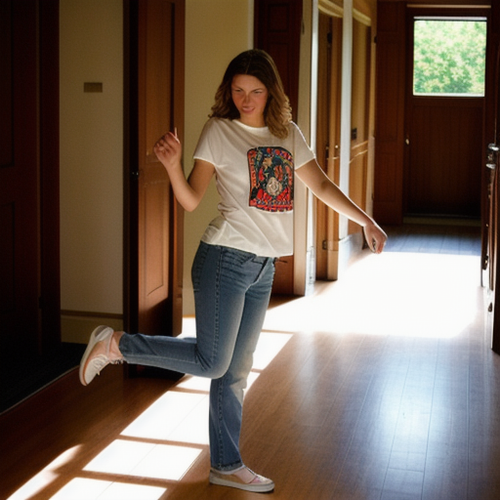} &%
\includegraphics[width=0.18\linewidth]{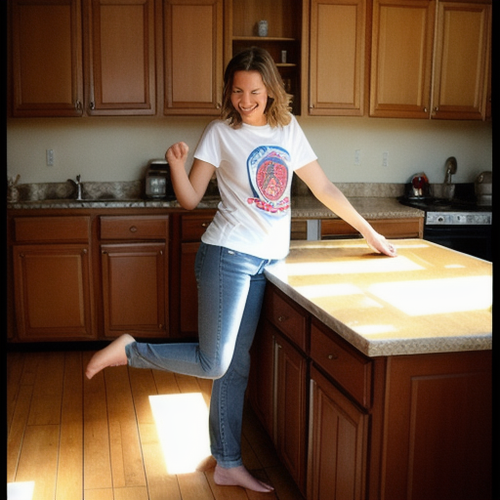} &%
\includegraphics[width=0.18\linewidth]{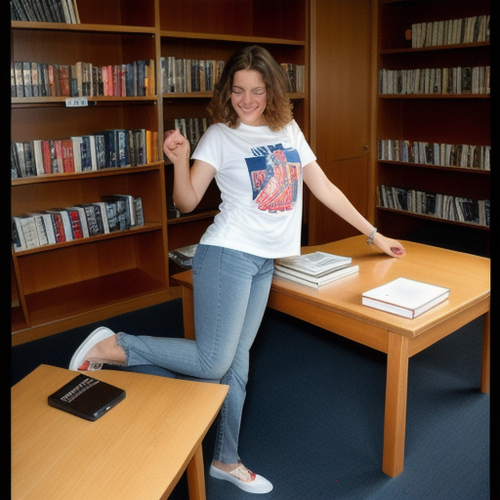} &%
\includegraphics[width=0.18\linewidth]{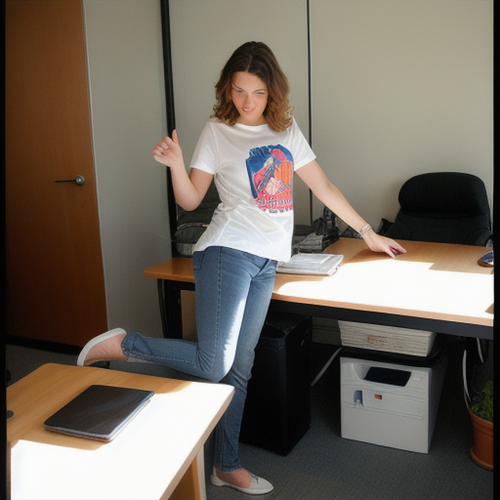} %
\\[-2pt]
& Hallway & Kitchen & Library & Office  \\[12pt]
\includegraphics[width=0.18\linewidth]{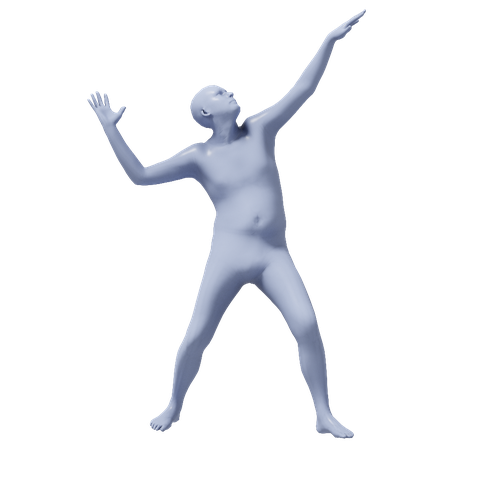} &%
\includegraphics[width=0.18\linewidth]{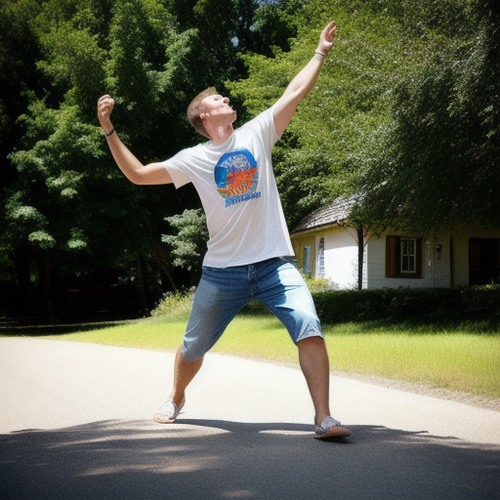} &%
\includegraphics[width=0.18\linewidth]{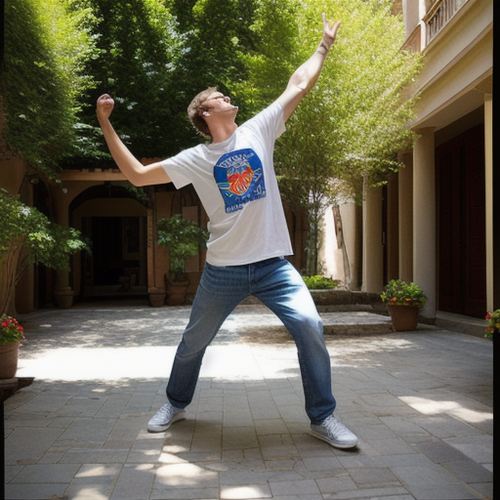} &%
\includegraphics[width=0.18\linewidth]{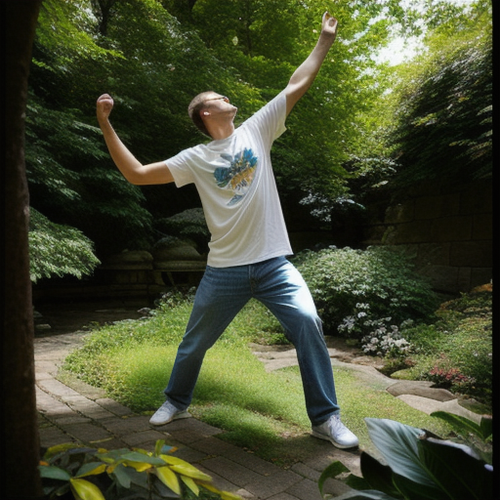} &%
\includegraphics[width=0.18\linewidth]{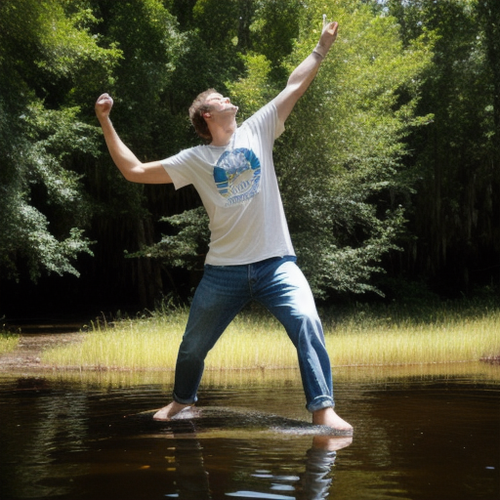} %
\\[-3pt]
& Village & Courtyard & Botanical garden & Swamp \\[12pt]
\includegraphics[width=0.18\linewidth]{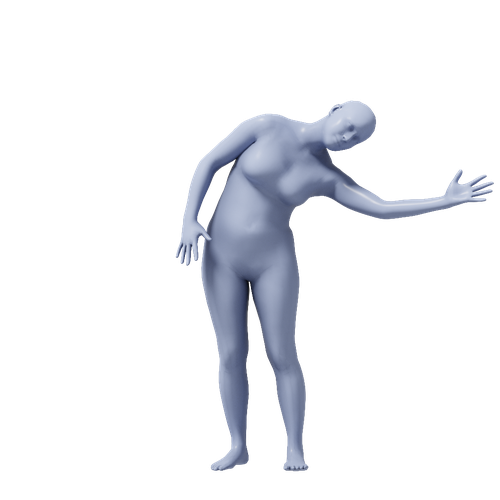} &%
\includegraphics[width=0.18\linewidth]{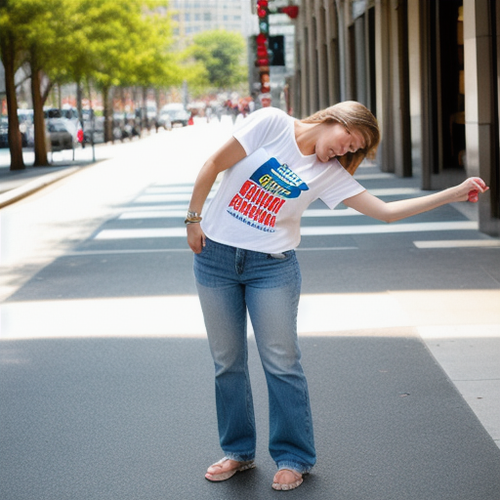} &%
\includegraphics[width=0.18\linewidth]{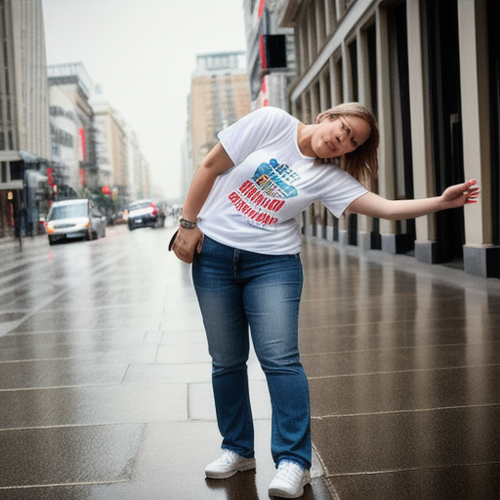} &%
\includegraphics[width=0.18\linewidth]{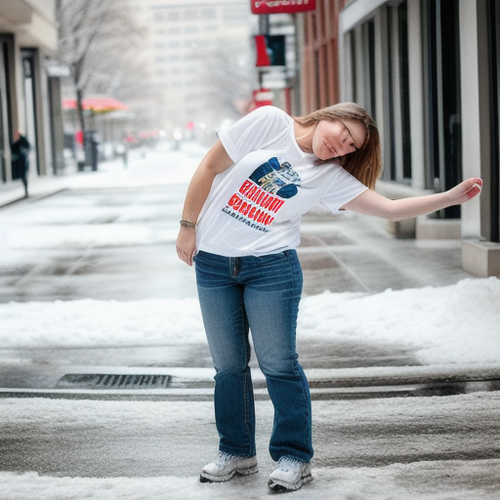} &%
\includegraphics[width=0.18\linewidth]{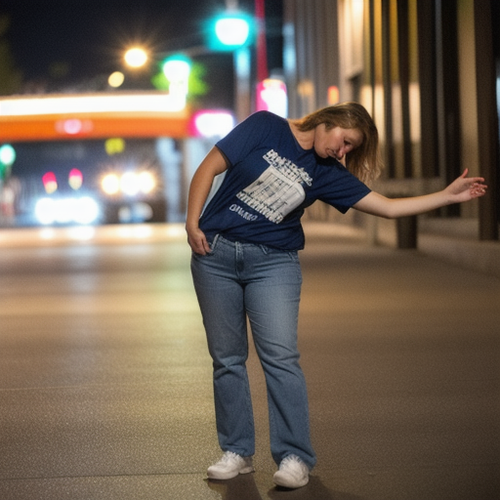} %
\\[-3pt]
& Sunny day & Rain & Snow & Night 
\end{tabular}
\par\end{centering}
\caption{
\textbf{Generated images for various attributes.} We present more examples of generated images. Based on a given ground truth pose (leftmost column), we generate images of people with different ethnicities, in different locations, or during different weather/lighting conditions. We start from a base prompt ``Photo, adult caucasian {male/female} wearing a t-shirt in the city center at daytime.'' and modify a single attribute, e.g., ``city center'' to ``gym''.
Images in the same row use the same initial noise for generation.
}%
\label{fig:supp:gallery}
\end{figure*}

The full results of our attribute experiments using STAGE are shown in \cref{fig:supp:clothing,fig:supp:texture,fig:supp:location:in,fig:supp:location:out,fig:supp:fairness,fig:supp:weather}.
We also provide more visual examples in \cref{fig:supp:gallery}, showing the utility of STAGE to test pose estimators with many different attributes.

We observe that none of the pose estimators are completely robust against attribute changes and can change their prediction if one aspect of the image is changed.
This is best observed in \cref{fig:supp:clothing,fig:supp:location:in}, where we see a continual increase of PDP  across all estimators from left to right.
This indicates that the more difficult attributes affect even the most robust models.

In the outdoor location experiments each location shows a similar impact, suggesting that they are equally challenging.
Only \promptstring{swamp} and \promptstring{wetlands} show a slightly higher impact compared to the rest of the attributes.
In contrast, the indoor locations differ from each other to a larger extent.
The most difficult indoor locations (\promptstring{bar}, \promptstring{restaurant} or \promptstring{kitchen}) are those where occlusions by tables are common.
Indeed, we observe that many such images depict occluded limbs, which suggests the leading cause of error for these attributes to be occlusions.

\section{Qualitative results}

\begin{figure*}%
\vspace{-1em}
\begin{centering}
\begin{tabular}{cccc}
\pbox{0.18\linewidth}{\centering CN-Pose} & \pbox{0.18\linewidth}{\centering CN-Depth} & CN-Multi  & CN-3DPose (ours) \\
\includegraphics[width=0.23\linewidth]{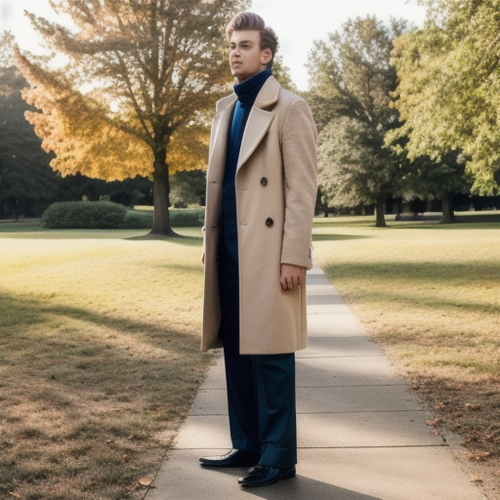} &%
\includegraphics[width=0.23\linewidth]{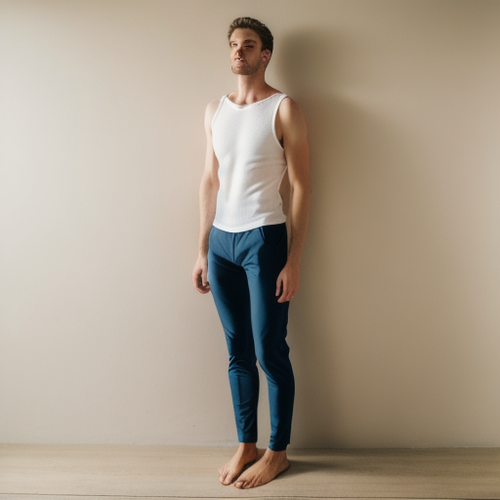} &%
\includegraphics[width=0.23\linewidth]{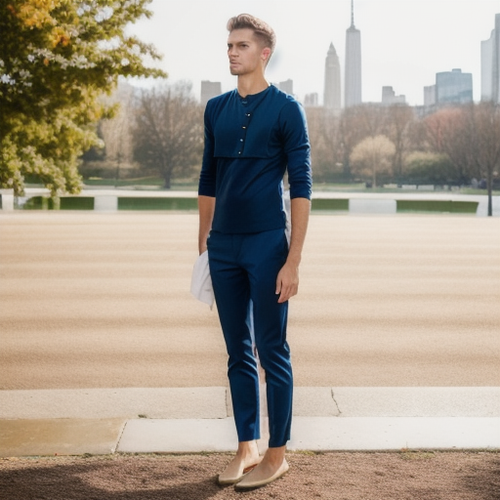} &%
\includegraphics[width=0.23\linewidth]{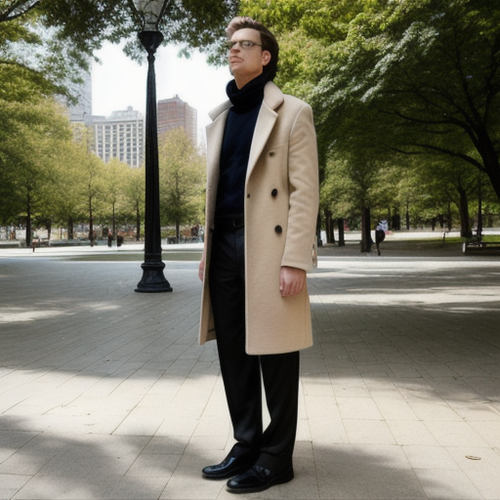} %
\\
\multicolumn{4}{c}{Photo, adult caucasian male wearing \textbf{long coat} in \textbf{city park} at daytime} \\[15pt]
\includegraphics[width=0.23\linewidth]{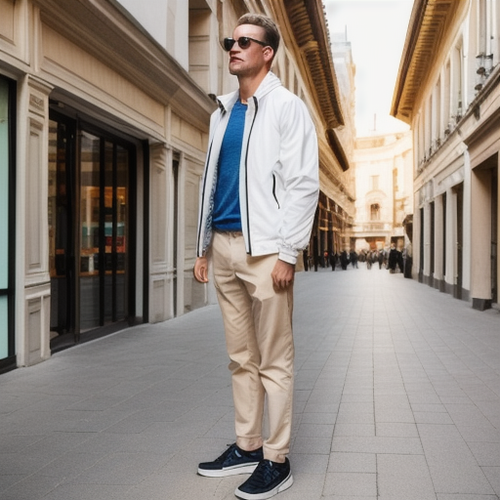} &%
\includegraphics[width=0.23\linewidth]{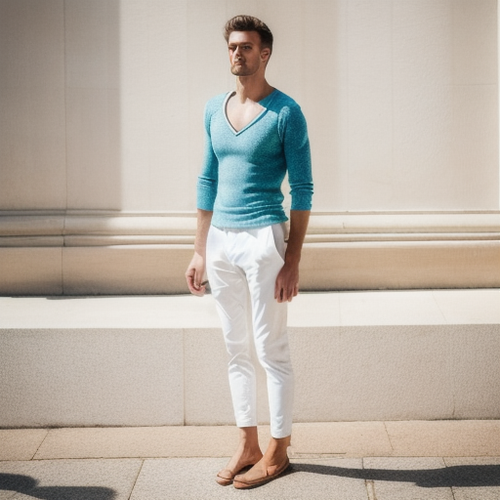} &%
\includegraphics[width=0.23\linewidth]{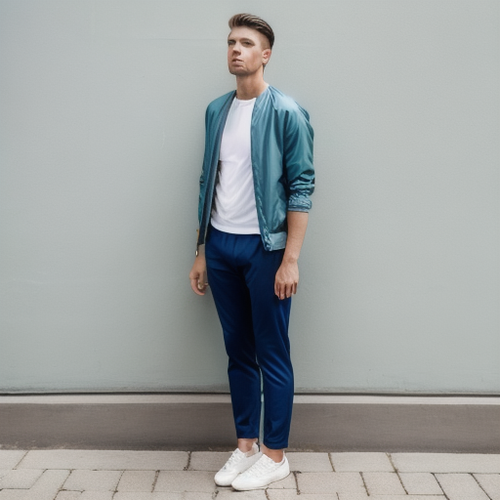} &%
\includegraphics[width=0.23\linewidth]{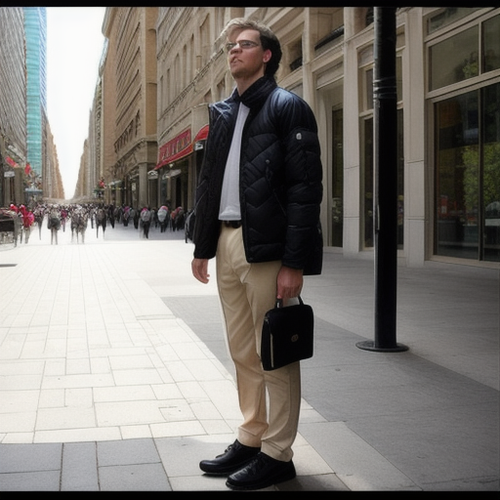} %
\\
\multicolumn{4}{c}{Photo, adult caucasian male wearing \textbf{jacket} in \textbf{city center} at daytime} \\[15pt]
\includegraphics[width=0.23\linewidth]{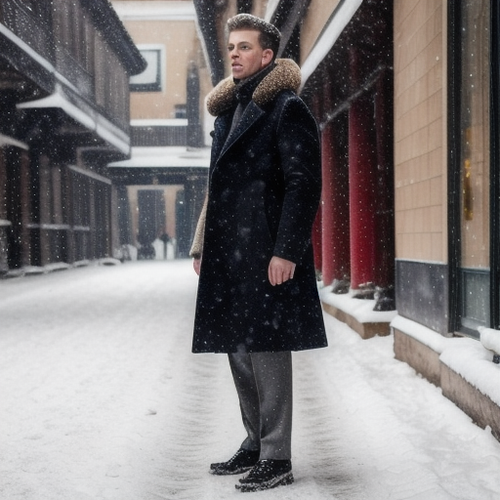} &%
\includegraphics[width=0.23\linewidth]{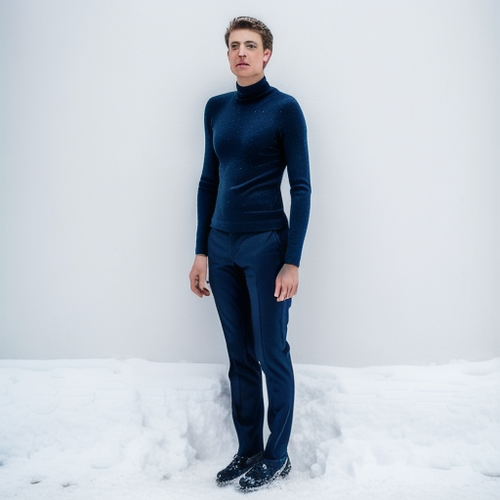} &%
\includegraphics[width=0.23\linewidth]{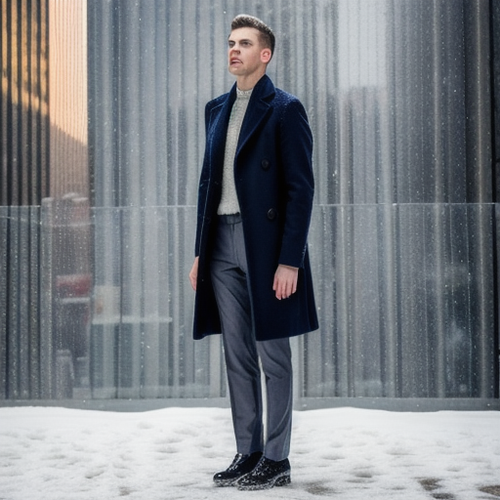} &%
\includegraphics[width=0.23\linewidth]{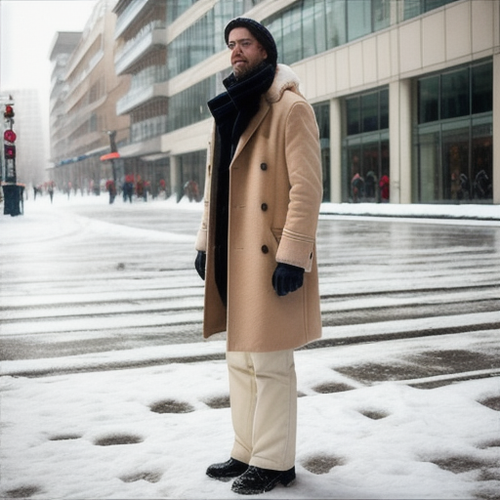} %
\\
\multicolumn{4}{c}{Photo, adult caucasian male wearing \textbf{long coat} in \textbf{city center} during snow} \\[15pt]
\includegraphics[width=0.23\linewidth]{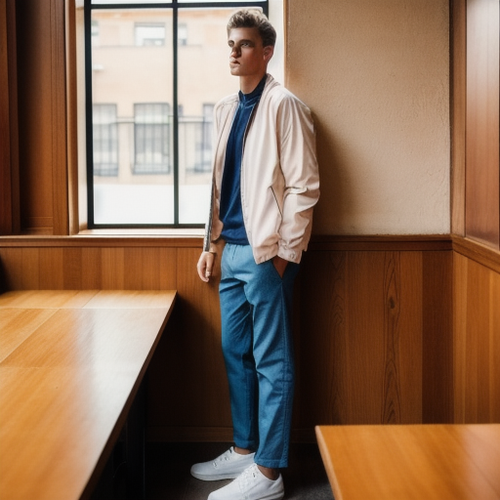} &%
\includegraphics[width=0.23\linewidth]{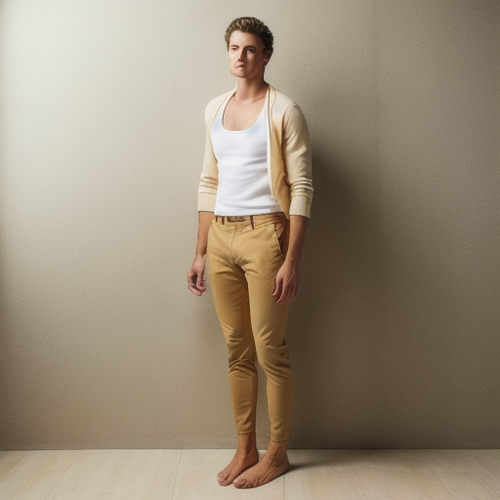} &%
\includegraphics[width=0.23\linewidth]{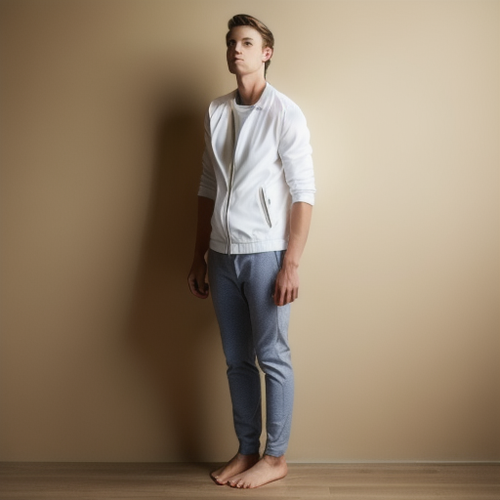} &%
\includegraphics[width=0.23\linewidth]{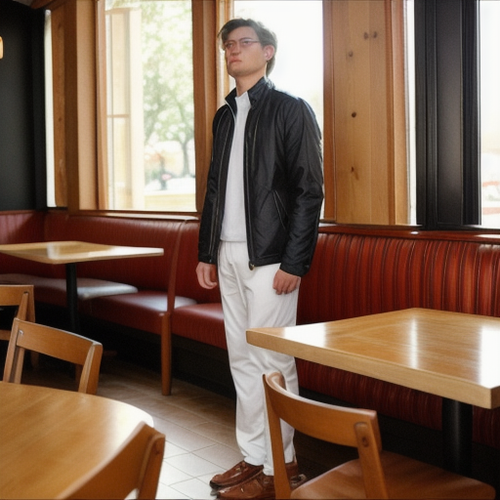} %
\\
\multicolumn{4}{c}{Photo, adult caucasian male wearing \textbf{jacket} in \textbf{restaurant} at daytime} \\
\end{tabular}
\par\end{centering}
\caption{
\textbf{Our method CN-3DPose generates diverse images.}
Given a single pose and multiple prompts we generate images with CN-Pose, CN-Depth, CN-Multi and CN-3DPose. CN-Depth and CN-Multi fail to follow the prompt and generate flat backgrounds or the wrong clothing item (notice the absence of a coat for CN-Depth and CN-Multi in row 1). Each image is generated from a different randomly sampled noise.
}%
\label{fig:supp:oursvs:2}
\end{figure*}

We compare our method CN-3DPose with CN-Pose, CN-Depth, and CN-Multi in terms of diversity in \cref{fig:supp:oursvs:2} by generating images with fixed input pose (per figure) and different prompts. 
CN-Depth and CN-Multi achieve good pose alignment but are not able to follow the prompt and generate only flat backgrounds.
CN-Pose creates diverse images but does not provide good pose alignment. 
Overall, only our method, CN-3DPose, can generate diverse images while having good pose alignment.

\end{document}